\documentclass[journal]{IEEEtai}

\usepackage[colorlinks,urlcolor=blue,linkcolor=blue,citecolor=blue]{hyperref}

\usepackage{color,array}
\usepackage{amsmath, amssymb}
\usepackage{graphicx}
\usepackage{booktabs}
\usepackage{algorithm, algorithmicx, algpseudocode}
\usepackage{comment}
\usepackage{subcaption} 

\newtheorem{theorem}{Theorem}

\newtheorem{remark}{Remark}
\newcommand{\rnn}{\texttt{RNN}}
\newcommand{\lstm}{\texttt{LSTM}}
\newcommand{\tcn}{\texttt{TCN}}
\newcommand{\ml}{\texttt{ML}}
\newcommand{\dl}{\texttt{DL}}
\newcommand{\cv}{\texttt{CV}}
\newcommand{\dnl}{\texttt{DL}}
\newcommand{\autoformer}{\texttt{Autoformer}}
\newcommand{\informer}{\texttt{Informer}}
\newcommand{\patchtst}{\texttt{PatchTST}}
\newcommand{\arima}{\texttt{ARIMA}}
\newcommand{\probsparse}{\texttt{ProbSparse}}
\newcommand{\nlp}{\texttt{NLP}}
\newcommand{\fedformer}{\texttt{FEDformer}}
\newcommand{\pyraformer}{\texttt{Pyraformer}}
\newcommand{\vit}{\texttt{ViT}}
\newcommand{\cnn}{\texttt{CNN}}

\begin{document}

\title{Synthetic Time Series Forecasting with Transformer Architectures: Extensive Simulation Benchmarks}

\author{Ali Forootani, \IEEEmembership{Senior, IEEE}, Mohammad Khosravi, \IEEEmembership{Member, IEEE}
\thanks{Ali Forootani is with Helmholtz Center for Environmental Research-UFZ, Permoserstraße 15, 04318 Leipzig, Germany, \texttt{Email: aliforootani@ieee.org/ali.forootani@ufz.de}.}
\thanks{
Mohammad Khosravi is with Delft Center for Systems and Control, Mekelweg 2, Delft, 2628 CD, The Netherlands, \texttt{Email: mohammad.khosravi@tudelft.nl}
}
}


\maketitle

\begin{abstract}

Time series forecasting plays a critical role in domains such as energy, finance, and healthcare, where accurate predictions inform decision-making under uncertainty. Although Transformer-based models have demonstrated success in sequential modeling, their adoption for time series remains limited by challenges such as noise sensitivity, long-range dependencies, and a lack of inductive bias for temporal structure. In this work, we present a unified and principled framework for benchmarking three prominent Transformer forecasting architectures—\autoformer, \informer, and \patchtst—each evaluated through three architectural variants: \textit{Minimal}, \textit{Standard}, and \textit{Full}, representing increasing levels of complexity and modeling capacity.

We conduct over 1500 controlled experiments on a suite of ten synthetic signals, spanning five patch lengths and five forecast horizons under both clean and noisy conditions. Our analysis reveals consistent patterns across model families. 

To advance this landscape further, we introduce the Koopman-enhanced Transformer framework, \textit{Deep Koopformer}, which integrates operator-theoretic latent state modeling to improve stability and interpretability. We demonstrate its efficacy on nonlinear and chaotic dynamical systems. Our results highlight Koopman based Transformer as a promising hybrid approach for robust, interpretable, and theoretically grounded time series forecasting in noisy and complex real-world conditions.

\end{abstract}



\section{Introduction}

\IEEEPARstart{T}{ime} series forecasting is a fundamental task in various domains including energy systems \cite{forootani2024climate, alvarez2010energy}, finance \cite{dingli2017financial}, supply chain management \cite{wang2024unveiling}, healthcare \cite{yan2024multi}, meteorology \cite{7438940}, and more. Accurate forecasting enables proactive decision-making, risk mitigation, and optimized planning. With the increasing availability of high-resolution temporal data, designing robust and scalable forecasting models has become more critical than ever. Traditional statistical methods such as Autoregressive Integrated Moving Average (\arima), Exponential Smoothing, and Seasonal Decomposition have long served as the backbone of time series analysis \cite{benidis2022deep}. However, their capacity to model complex patterns, multivariate dependencies, and non-linear dynamics is inherently limited, particularly in large-scale or non-stationary datasets.

In recent years, machine learning (\ml) and deep learning (\dnl) models have emerged as powerful alternatives to classical methods due to their data-driven learning capacity and flexibility. Recurrent Neural Networks (\rnn s), Long Short-Term Memory (\lstm) networks, and Temporal Convolutional Networks (\tcn s) have been successfully applied to time series forecasting tasks, demonstrating superior performance in capturing temporal dependencies and non-linear patterns \cite{hewage2020temporal}. Nevertheless, these models often suffer from challenges such as vanishing gradients, sequential bottlenecks, and limited scalability \cite{survey1,survey2,survey3}.

Among recent advancements in \dl, Transformer models have revolutionized sequence modeling, particularly in Natural Language Processing (\nlp) \cite{nlpsurvey}. Introduced by \cite{transformer}, the Transformer architecture replaces recurrence with self-attention mechanisms, enabling parallel computation and long-range dependency modeling. Its success has quickly spread to other domains such as Computer Vision (\cv) \cite{cvsurvey}, speech processing \cite{speechsurvey}, and more recently, time series forecasting \cite{tssurvey}. Transformers provide an appealing framework for time series modeling due to their flexibility in handling variable-length inputs, modeling contextual dependencies, and learning rich representations \cite{fournier2023practical}.

However, applying Transformers directly to time series forecasting is not straightforward. Time series data differ from text or images in several key aspects. First, time series are continuous and regularly sampled, whereas text and vision data have discrete semantic tokens (e.g., words, pixels). Second, time series often exhibit strong autocorrelation, trends, and seasonalities—features that need to be explicitly modeled. Third, computational efficiency is a major concern, especially for long-term forecasting where the input sequence can be significantly large. The standard Transformer has quadratic complexity $\mathcal{O}(N^2)$ with respect to the sequence length $N$, making it computationally expensive for high-resolution temporal data \cite{zhu2021long}.

To overcome these challenges, a variety of Transformer-based architectures have been proposed for time series forecasting. \informer~\cite{informer} introduces \probsparse~attention to select important queries and reduce redundancy in self-attention computation. \autoformer~ \cite{autoformer} incorporates a decomposition block to handle trends and seasonality explicitly, while \fedformer~\cite{fedformer} integrates Fourier-based decomposition to reduce complexity. \pyraformer~\cite{pyraformer} utilizes a pyramidal attention mechanism to capture hierarchical temporal features. Despite their differences, these models share a common goal: reducing computational cost while preserving or enhancing predictive accuracy.

Nevertheless, a recent study by \cite{dlinear} challenges the assumption that complex Transformer models are necessary for high-accuracy time series forecasting. The authors propose a simple linear model, \texttt{DLinear}, which surprisingly outperforms many sophisticated models on standard benchmarks. This raises fundamental questions about the design of deep models for time series and whether architectural complexity always translates to improved performance. In \cite{nie2022time}, the authors introduced \patchtst~to enhance the model's ability to capture local and semantic dependencies through patching, and to simplify the design by leveraging \emph{channel-independent} learning.

Unlike traditional Transformer variants that treat each time step as a token, \patchtst~argues that aggregating multiple time steps into local patches better reflects the temporal structure of time series data. This approach is inspired by techniques in Vision Transformers (\vit) \cite{vit} and speech models such as Wav2Vec 2.0 \cite{wav2vec2}, where patches or segments are used to encode meaningful context.

Furthermore, in multivariate time series, the common practice is to mix channels—i.e., combine features from all variables into a single token. While this enables joint learning, it may obscure individual patterns and increase overfitting risks. In \patchtst~a \emph{channel-independent} formulation has been adapted, where each channel (i.e., time series variable) is treated separately during embedding and encoding. This strategy is shown to work well in Convolutional Neural Network (\cnn) and linear models \cite{multichannel,dlinear}, and has been extended in \patchtst~ for Transformer-based models.

Even though Transformer-based architectures have revolutionized sequential modeling in natural language processing and computer vision, yet their adoption in time series forecasting remains underdeveloped. Unlike textual data, time series are often governed by latent temporal dynamics, periodic structures, and long-range dependencies that challenge standard attention mechanisms. Moreover, existing Transformer variants such as \autoformer, \informer, and \patchtst~have been proposed independently, each introducing unique architectural innovations—but without a unified framework for comparison. This lack of standardization hinders practical model selection, reproducibility, and theoretical understanding. To address this gap, we conduct a systematic study of these models, introduce consistent architectural variants, and evaluate them under controlled experimental settings. Our goal is to clarify the design trade-offs, computational properties, and forecasting behaviors of Transformer models, ultimately guiding practitioners toward more informed and effective forecasting solutions.

\noindent (\textbf{i}) \textit{Architectural Variant Framework.} We design and evaluate three principled variant classes—\textit{Minimal}, \textit{Standard}, and \textit{Full}—for each of the \autoformer, \informer, and \patchtst~families. These variants represent increasing architectural complexity, ranging from lightweight encoder-only models to full encoder-decoder configurations with autoregressive decoding. This design space allows us to isolate the impact of architectural choices on performance, scalability, and robustness.

\noindent (\textbf{ii}) \textit{Unified Forecasting and Complexity Formulation.} All variants are analyzed within a shared theoretical framework, encompassing patch-based input encoding, trend-seasonal decomposition (\autoformer), ProbSparse attention (\informer), and temporal patch tokenization (\patchtst). We provide consistent mathematical formulations, computational complexity comparisons, and layer-by-layer forward pass descriptions that contextualize how each model processes temporal inputs.

\noindent (\textbf{iii}) \textit{Comprehensive Empirical Benchmarking under Noise.} We conduct 1500 controlled experiments (750 clean + 750 noisy) across 10 synthetic signals, 5 patch lengths, and 5 forecast horizons for each model family. Our analysis reveals clear and consistent trends: \patchtst~Standard achieves the best overall performance under clean and noisy conditions; \autoformer~Minimal and \autoformer~Standard excel on smooth and trend-dominated signals, especially in noisy regimes; while \informer~variants exhibit higher error and instability under noise and long-horizon forecasting. Aggregated heatmaps and per-signal tables highlight the trade-offs between robustness, sensitivity, and generalization across the architectural spectrum.

\noindent (\textbf{iv}) \textit{Koopman-Enhanced Transformer Architecture.} We introduce a novel extension of Transformer-based forecasting models that integrates Koopman operator theory to model linear latent state evolution. This modular operator-theoretic approach improves forecast stability, enables interpretable latent dynamics, and bridges deep sequence modeling with dynamical systems analysis. Through experiments on nonlinear and chaotic systems, including noisy Van der Pol and Lorenz systems, we demonstrate Koopman enhanced Transformer’s capacity to effectively forecast nonlinear oscillations and chaotic trajectories under stochastic perturbations.


This paper is organized as follows: In the Section \ref{preliminaries_sec}, we review preliminaries and notations that we use in this article. In Sections \ref{patch_sec}, \ref{informer_sec}, and \ref{autoformer_sec} we propose three different variants for transformer based time series forecasting, i.e. \patchtst, \informer, and \autoformer~, respectively. In Section \ref{simulation_sec} we evaluate the performance of these architectures on 10 clean and noisy synthetic signals. The koopman enhanced transformer architectures will be discussed in Section \ref{koopman}. Finally, the conclusion will be discussed in Section \ref{conclusion_sec}.

\section{Preliminaries and Notation}\label{preliminaries_sec}

We begin by formalizing the forecasting setup and introducing common notation and operations used throughout this work.

\paragraph{Time Series Data}
Let $x = [x_1, x_2, \dots, x_T] \in \mathbb{R}^T$ denote a univariate time series of length $T$, where $x_t \in \mathbb{R}$ represents the observation at time step $t$. Our objective is to predict future values of the sequence based on historical observations.

\paragraph{Forecasting Task}
Given a historical context window of fixed length $P$, the model forecasts the next $H$ steps. For each starting index $i \in \{1, 2, \dots, T - P - H + 1\}$, we define:
\begin{align}
\mathbf{x}_i &= [x_i, x_{i+1}, \dots, x_{i+P-1}] \in \mathbb{R}^{P}, \\
\mathbf{y}_i &= [x_{i+P}, x_{i+P+1}, \dots, x_{i+P+H-1}] \in \mathbb{R}^{H}.
\end{align}
The goal is to learn a forecasting function $f_\theta: \mathbb{R}^P \rightarrow \mathbb{R}^H$ that maps the historical window to future values:
\begin{equation}
\hat{\mathbf{y}}_i = f_\theta(\mathbf{x}_i).
\end{equation}

\paragraph{Common Operations and Notation}
Throughout the article, we use the following notation:

\begin{itemize}
    \item \emph{Input Embedding}: Each input $\mathbf{x}_i$ is mapped to a latent representation via a linear embedding:
    \[
    \mathbf{z}_i = \mathbf{x}_i \cdot \mathbf{W}_e, \quad \mathbf{W}_e \in \mathbb{R}^{P \times d_{\text{model}}} \text{ or } \mathbb{R}^{1 \times d_{\text{model}}}.
    \]
    
    \item \emph{Positional Encoding (PE)}: To inject temporal order, a positional encoding $\text{PE} \in \mathbb{R}^{P \times d_{\text{model}}}$ is added:
    \[
    \mathbf{z}_i^{\text{pos}} = \mathbf{z}_i + \text{PE},
    \]
    where PE is either:
    \begin{itemize}
        \item \emph{Fixed sinusoidal}: \(\text{PE}_{(pos,2k)} = \sin\left(\frac{pos}{10000^{2k/d_{\text{model}}}}\right)\), \(\text{PE}_{(pos,2k+1)} = \cos\left(\frac{pos}{10000^{2k/d_{\text{model}}}}\right)\),
        \item or \textit{Learnable}: parameters optimized during training.
    \end{itemize}
    
    \item \emph{Transformer Encoder}: Given a sequence $\mathbf{Z} \in \mathbb{R}^{P \times d_{\text{model}}}$, the Transformer encoder applies multiple layers of multi-head self-attention (MSA) and feedforward networks (FFN):
    \begin{multline}
         \mathbf{H}^{(e)}=\text{TransformerEncoder}(\mathbf{Z})\\ = \text{LayerNorm}\left( \mathbf{Z} + \text{FFN}\left(\text{MSA}(\mathbf{Z})\right) \right),
    \end{multline}
    where $\mathbf{H}^{(e)} \in \mathbb{R}^{P \times d_{\text{model}}}$.
    \item \text{Transformer Decoder}: Given a decoder input $\mathbf{Z}^{(d)}$ and encoder output $\mathbf{H}^{(e)}$, the Transformer decoder applies self-attention over $\mathbf{Z}^{(d)}$ and cross-attention with $\mathbf{H}^{(e)}$:
    \begin{multline}
     \mathbf{H}^{(d)} = \text{TransformerDecoder}(\mathbf{Z}^{(d)}, \mathbf{H}^{(e)}) \\= \text{LayerNorm}\Bigg( \mathbf{Z}^{(d)} \\+ \text{CrossAttention}\left(\text{SelfAttention}(\mathbf{Z}^{(d)}), \mathbf{H}^{(e)}\right) \Bigg).
    \end{multline}
    
    \item \emph{Average Pooling (AvgPool)}: After encoding, we often apply average pooling over the sequence (token) dimension to get a fixed-size vector:
    \[
    \text{AvgPool}(\mathbf{H}) = \frac{1}{P} \sum_{p=1}^{P} \mathbf{H}_{p},
    \]
    where $\mathbf{H}_{p}$ denotes the $p$-th token embedding.
    
    \item \emph{Output Projection}: The final output is generated by projecting the pooled or decoded representation:
    \[
    \hat{\mathbf{y}}_i = \mathbf{W}_o \cdot \mathbf{h}_i,
    \]
    where $\mathbf{W}_o \in \mathbb{R}^{d_{\text{model}} \times H}$ and $\mathbf{h}_i$ is the pooled hidden state.
\end{itemize}

In the \autoformer~model, the time series is decomposed into trend and seasonal components. The trend component, \(\mathbf{x}^{\text{trend}}\), is obtained by applying a moving average (MA) filter with a kernel size \(k\). The kernel size \(k\) defines the window over which the average is computed. This filter smooths the time series to capture long-term trends, while the residual seasonal component, \(\mathbf{x}^{\text{seasonal}}\), captures short-term periodic fluctuations.

In \informer, the attention mechanism is enhanced using a sparse attention approach to improve efficiency. The sparsity score \(M(q_i)\) for each query \(q_i\) is computed by first calculating the attention between each query and all keys, and then applying a softmax function to identify the most relevant keys for each query. The sparsity score \(M(q_i)\) is the difference between the maximum attention value and the average attention value over all keys, ensuring that only the most important keys are selected for attention computation. This enables the model to focus on a smaller subset of keys, significantly reducing computational complexity while retaining high performance.





\section{\patchtst~Variants: Architecture and Computational Complexity}\label{patch_sec}

The \patchtst~framework addresses accurate forecasting of univariate time series by first segmenting the input sequence into overlapping or non-overlapping patches of fixed-length patches, then modeling temporal dependencies using a Transformer-based architecture. In this section, we present and justify three architectural variants—\textit{Minimal}, \textit{Standard}, and \textit{Full}—each representing a progressively more flexible and expressive model design.

\subsection{\patchtst~Minimal: A Lightweight Transformer Backbone}
The Minimal variant serves as a baseline design, emphasizing simplicity and computational efficiency. Each input patch $\mathbf{x}_i$ is linearly projected into a higher-dimensional latent space:
\begin{equation}
\mathbf{z}_i = \mathbf{x}_i \cdot \mathbf{W}_e \in \mathbb{R}^{1 \times d_{\text{model}}},
\end{equation}
Fixed sinusoidal positional encodings are added to incorporate temporal information:
\begin{equation}
\mathbf{z}_i^{\text{pos}} = \mathbf{z}_i + \text{PE},
\end{equation}
This encoded representation is passed through a Transformer encoder, followed by average pooling and a linear projection:
\begin{align}
\mathbf{h}_i &= \text{TransformerEncoder}(\mathbf{z}_i^{\text{pos}}), \\
\hat{\mathbf{y}}_i &= \mathbf{W}_o \cdot \text{AvgPool}(\mathbf{h}_i) \in \mathbb{R}^H,
\end{align}
This variant is lightweight and suitable for short sequences or fast, interpretable inference.
\paragraph{Computational Complexity}
\[
\mathcal{O}_{\text{Minimal}} = 
\underbrace{P \cdot d_{\text{model}}^2}_{\text{Feedforward}} + 
\underbrace{P^2 \cdot d_{\text{model}}}_{\text{Self-Attention}}.
\]

\subsection{\patchtst~Standard: Learning Data-Driven Positional Embeddings}
The Standard variant replaces the fixed positional encodings with a learnable global positional embedding $\mathbf{p} \in \mathbb{R}^{1 \times d_{\text{model}}}$:
\begin{equation}
\mathbf{z}_i^{\text{pos}} = \mathbf{x}_i \cdot \mathbf{W}_e + \mathbf{p},
\end{equation}
This embedding is shared across all positions and learned during training, allowing the model to adapt its temporal representation. The rest of the architecture remains identical to Minimal architecture:
\[
\hat{\mathbf{y}}_i = \mathbf{W}_o \cdot \text{AvgPool}(\text{TransformerEncoder}(\mathbf{z}_i^{\text{pos}})).
\]

\paragraph{Computational Complexity}
\begin{equation*}
\mathcal{O}_{\text{Standard}} = \mathcal{O}_{\text{Minimal}} + 
\underbrace{\mathcal{O}(d_{\text{model}})}_{\text{Embedding overhead (negligible)}}.  
\end{equation*}

\subsection{\patchtst~Full: A Flexible Encoder-Decoder Architecture}

The Full variant introduces a decoder with cross-attention, allowing the model to explicitly learn dependencies between past inputs and future predictions.

\paragraph{Encoder} The input patch is embedded and encoded with sinusoidal positional encodings:
\begin{align}
\mathbf{Z}_i^{(e)} &= \mathbf{x}_i \cdot \mathbf{W}_e^{(e)} + \text{PE}^{(e)} \in \mathbb{R}^{P \times d_{\text{model}}}, \\
\mathbf{H}_i^{(e)} &= \text{TransformerEncoder}(\mathbf{Z}_i^{(e)}).
\end{align}

\paragraph{Decoder} The decoder input is formed by repeating the last element of the patch to enable autoregressive-style conditioning:
\begin{align}
\mathbf{x}_i^{\text{rep}} &= [x_{i+P-1}, \dots, x_{i+P-1}] \in \mathbb{R}^{H}, \\
\mathbf{Z}_i^{(d)} &= \mathbf{x}_i^{\text{rep}} \cdot \mathbf{W}_e^{(d)} + \text{PE}^{(d)} \in \mathbb{R}^{H \times d_{\text{model}}}, \\
\mathbf{H}_i^{(d)} &= \text{TransformerDecoder}(\mathbf{Z}_i^{(d)}, \mathbf{H}_i^{(e)}).
\end{align}

\paragraph{Output} The decoder output is projected to obtain the forecast:

\begin{equation}
\hat{\mathbf{y}}_i = \mathbf{W}_o \cdot \mathbf{H}_i^{(d)} \in \mathbb{R}^{H}.
\end{equation}

\paragraph{Computational Complexity}
\[
\mathcal{O}_{\text{Full}} = 
\underbrace{P \cdot d_{\text{model}}^2}_{\text{Encoder FFN}} +
\underbrace{P^2 \cdot d_{\text{model}}}_{\text{Encoder Attention}} +
\underbrace{H \cdot d_{\text{model}}^2}_{\text{Decoder FFN}} +
\underbrace{H \cdot P \cdot d_{\text{model}}}_{\text{Cross-Attention}}.
\]

\subsection{Comparative Summary}
Each \patchtst~variant is designed to address different forecasting needs and trade-offs: (i) The \textit{Minimal} variant provides fast training and interpretable forecasts, well-suited for small datasets and short horizons; (ii) The \textit{Standard} variant adds learnable temporal representations, improving adaptability to real-world signals without significant computational overhead; (iii) The \textit{Full} variant introduces an encoder-decoder structure that improves long-horizon modeling at the cost of increased complexity. A comparative overview of \patchtst~variants is shown in Table~\ref{tab:patchtst_variants}, and the complete forecasting procedure for the Full variant is provided in Algorithm~\ref{alg:patchtstfull} in Appendix.

\begin{table*}[h]
\renewcommand{\arraystretch}{1.3}
\caption{Comparison of \patchtst~ Variants}
\label{tab:patchtst_variants}
\centering
\begin{tabular}{|c|c|c|c|}
\hline
Model Variant & Positional Encoding & Architecture     & Time Complexity \\
\hline
Minimal \patchtst  & Sinusoidal          & Encoder-only     & $\mathcal{O}(P^2 d_{\text{model}} + P d_{\text{model}}^2)$ \\
Standard \patchtst & Trainable (global)  & Encoder-only     & $\mathcal{O}(P^2 d_{\text{model}} + P d_{\text{model}}^2)$ \\
Full \patchtst & Sinusoidal (dual)   & Encoder-Decoder  & $\mathcal{O}(P^2 d_{\text{model}} + H P d_{\text{model}} + H d_{\text{model}}^2)$ \\
\hline
\end{tabular}
\end{table*}


\section{\informer~ Variants: Architecture and Computational Complexity}\label{informer_sec}

Transformer models have demonstrated strong performance in sequence modeling, yet their standard self-attention mechanism suffers from quadratic time complexity with respect to input length. This poses a bottleneck for long time series forecasting. To overcome this, we present three variants based on the \informer~framework—Minimal, Standard, and Full—each offering a different balance between expressiveness and computational efficiency.

\subsection{\informer~Minimal: Baseline with Full Self-Attention}
The Minimal variant adopts a standard encoder-only Transformer architecture. Each input patch is projected into a latent space using a linear transformation:
\begin{equation}
\mathbf{z}_i = \mathbf{x}_i \cdot \mathbf{W}_e \in \mathbb{R}^{P \times d_{\text{model}}}.
\end{equation}

To retain temporal information, fixed sinusoidal positional encodings are added:
\begin{equation}
\mathbf{z}_i^{\text{pos}} = \mathbf{z}_i + \text{PE}(i),
\end{equation}
This sequence is passed through a multi-layer Transformer encoder:
\begin{align}
\mathbf{h}_i &= \text{TransformerEncoder}(\mathbf{z}_i^{\text{pos}}),\\
\hat{\mathbf{y}}_i &= \mathbf{W}_o \cdot \text{AvgPool}(\mathbf{h}_i),
\end{align}
This model serves as a full-attention baseline, but its complexity increases quadratically with the input length.

\paragraph{Computational Complexity}
\[
\mathcal{O}_{\text{Minimal}} = 
\underbrace{P^2 \cdot d_{\text{model}}}_{\text{Self-Attention}} +
\underbrace{P \cdot d_{\text{model}}^2}_{\text{Feedforward Layers}}
\]

\begin{remark}
The primary difference between \patchtst~Minimal and \informer~Minimal lies in how they process time series data and the underlying architectural philosophy they follow. \patchtst~Minimal adopts a patch-based approach where the input sequence is segmented into fixed-length patches, which are then embedded and processed by a Transformer encoder. This design focuses on capturing local patterns efficiently, reducing the input sequence length and computational complexity. In contrast, \informer~Minimal follows a more conventional Transformer structure, directly operating on the full-resolution time series using positional encoding and self-attention across all time steps. This enables it to retain fine-grained temporal dependencies but can be more computationally intensive for long sequences. While \patchtst~Minimal excels in modeling smooth, low-frequency patterns with fewer tokens, \informer~Minimal is better suited for capturing detailed, high-frequency dynamics in time series data.
\end{remark}

\subsection{\informer~Standard: ProbSparse Attention for Scalable Forecasting}
The Standard variant improves scalability by replacing full attention with a sparse approximation known as ProbSparse attention. This mechanism identifies a subset of ``informative'' queries and restricts attention to them.

\paragraph{Query Selection} For each query vector $\mathbf{q}_i$, a sparsity score is computed:

\begin{equation}
M(\mathbf{q}_i) = \max_j \left( \frac{\mathbf{q}_i \cdot \mathbf{k}_j}{\sqrt{d_{\text{model}}}} \right) - \frac{1}{P} \sum_{j=1}^{P} \frac{\mathbf{q}_i \cdot \mathbf{k}_j}{\sqrt{d_{\text{model}}}},
\end{equation}

Only the top-$u$ queries with the highest scores (where $u = \mathcal{O}(\log P)$) are selected to perform full attention over all keys.

\paragraph{Encoding} These selected queries are processed using the ProbSparse encoder:
\begin{align}
\mathbf{h}_i &= \text{ProbSparseEncoder}(\mathbf{z}_i^{\text{pos}}), \\
\hat{\mathbf{y}}_i &= \mathbf{W}_o \cdot \text{AvgPool}(\mathbf{h}_i).
\end{align}

\paragraph{Computational Complexity}
\[
\mathcal{O}_{\text{Standard}} = 
\underbrace{\log P \cdot P \cdot d_{\text{model}}}_{\text{Sparse Attention}} +
\underbrace{P \cdot d_{\text{model}}^2}_{\text{Feedforward Layers}}.
\]

This represents a significant improvement over the quadratic complexity of the Minimal variant.

\subsection{\informer~Full: Sequence-to-Sequence Architecture with Dual Attention}

The Full variant extends \informer~into a full encoder-decoder architecture. Both the encoder and decoder employ ProbSparse attention, enabling modeling of long-range dependencies in both input and output sequences.

\paragraph{Encoder}
\begin{align}
\mathbf{z}_i^{(e)} &= \mathbf{x}_i \cdot \mathbf{W}_e^{(e)} + \text{PE}^{(e)}, \\
\mathbf{H}_i^{(e)} &= \text{ProbSparseEncoder}(\mathbf{z}_i^{(e)}).
\end{align}

\paragraph{Decoder}
\begin{align}
\mathbf{x}_i^{\text{rep}} &= [x_{i+P-1}, \dots, x_{i+P-1}] \in \mathbb{R}^H, \\
\mathbf{z}_i^{(d)} &= \mathbf{x}_i^{\text{rep}} \cdot \mathbf{W}_e^{(d)} + \text{PE}^{(d)}, \\
\mathbf{H}_i^{(d)} &= \text{ProbSparseDecoder}(\mathbf{z}_i^{(d)}, \mathbf{H}_i^{(e)}).
\end{align}

\paragraph{Output}
\begin{equation}
\hat{\mathbf{y}}_i = \mathbf{W}_o \cdot \mathbf{H}_i^{(d)} \in \mathbb{R}^H.
\end{equation}

\paragraph{Computational Complexity}
\[
\mathcal{O}_{\text{Full}} = 
\underbrace{\log P \cdot P \cdot d_{\text{model}}}_{\text{Encoder Sparse Attention}} +
\underbrace{\log H \cdot H \cdot d_{\text{model}}}_{\text{Decoder Sparse Attention}} +
\underbrace{H \cdot P \cdot d_{\text{model}}}_{\text{Cross-Attention}}
\]

\subsection{Comparative Summary}

Each \informer~ variant offers different strengths: (i) \textit{Minimal} serves as a full-attention benchmark and is effective for short sequences, but computationally expensive; (ii) \textit{Standard} improves scalability using probabilistic attention, offering a good balance of efficiency and accuracy; (iii) \textit{Full} introduces a sequence-to-sequence structure for modeling long-term, non-stationary patterns, at higher computational cost. A comparative overview is shown in Table~\ref{tab:informer_variants}, and the complete forecasting procedure for the Full variant is provided in Algorithm~\ref{alg:fullinformer} in Appendix. The following Theorem we prove the reduction complexity resulted by employing ProbSparse technique.

\begin{table*}[!t]
\renewcommand{\arraystretch}{1.3}
\caption{Comparison of \informer~Variants}
\label{tab:informer_variants}
\centering
\begin{tabular}{|c|c|c|c|}
\hline
Model Variant & Attention Type         & Architecture     & Time Complexity \\
\hline
Minimal \informer& Full Self-Attention      & Encoder-only     & $\mathcal{O}(P^2 d_{\text{model}} + P d_{\text{model}}^2)$ \\
Standard \informer & ProbSparse (log-scaled)  & Encoder-only     & $\mathcal{O}(\log P \cdot P d_{\text{model}} + P d_{\text{model}}^2)$ \\
Full \informer & ProbSparse (dual)        & Encoder-Decoder  & $\mathcal{O}(\log P \cdot P d_{\text{model}} + \log H \cdot H d_{\text{model}} + H P d_{\text{model}})$ \\
\hline
\end{tabular}
\end{table*}



\begin{theorem}
(Attention Complexity Reduction in ProbSparse \informer)  
If the attention mechanism is dominated by a small number of significant query-key pairs, ProbSparse reduces the attention complexity from $\mathcal{O}(L_Q L_K)$ to $\mathcal{O}(L_Q \log L_K)$ with bounded approximation error.
\end{theorem}

\begin{proof}
Let $Q \in \mathbb{R}^{L_Q \times d_{\text{model}}}$ and $K \in \mathbb{R}^{L_K \times d_{\text{model}}}$ be the query and key matrices. For each query $q_i$, define a sparsity score:
\[
M(q_i, K) = \max_j \left( \frac{q_i k_j^\top}{\sqrt{d_{\text{model}}}} \right) - \frac{1}{L_K} \sum_{j=1}^{L_K} \frac{q_i k_j^\top}{\sqrt{d_{\text{model}}}}.
\]
Let $u = c \log L_K$ for some constant $c > 0$. Compute full attention for the top-$u$ queries: $\mathcal{O}(u L_K) = \mathcal{O}(L_K \log L_K)$. Approximate the remaining queries using attention over $u$ keys: $\mathcal{O}((L_Q - u) \cdot u) = \mathcal{O}(L_Q \log L_K)$.
\[
\text{Total cost: } \mathcal{O}(L_Q \log L_K), \quad \text{assuming } L_Q \sim L_K.
\]
The approximation error satisfies:
\[
\|\Delta \text{Attention}\| \leq C \epsilon,
\]
where $\epsilon$ decays exponentially with $u$, and $C$ is a constant. Thus, the reduction in complexity is achieved without significant loss in accuracy.
\end{proof}






\section{\autoformer~Variants: Architecture and Computational Complexity}\label{autoformer_sec}
\autoformer~addresses the limitations of Transformer-based time series forecasting by introducing a decomposition architecture that separates each input sequence into trend and seasonal components. This inductive bias improves forecasting stability and interpretability, especially for signals with periodic or long-term structural patterns. We present three architectural variants—Minimal, Standard, and Full—tailored to different forecasting tasks and computational budgets.

\autoformer~applies a moving average filter $\text{MA}_k$ of kernel size $k$ to decompose each input into trend and seasonal components:
\begin{align}
\mathbf{x}_i^{\text{trend}} &= \text{MA}_k(\mathbf{x}_i), \\
\mathbf{x}_i^{\text{seasonal}} &= \mathbf{x}_i - \mathbf{x}_i^{\text{trend}},
\end{align}
This decomposition is used throughout all model variants.

\subsection{\autoformer~Minimal: Lightweight Forecasting with Minimal Overhead}

The Minimal variant retains the decomposition structure but applies a short moving average (e.g. $k = 3$) and a shallow encoder-only Transformer. The seasonal component is embedded and positionally encoded:

\begin{align}
\mathbf{z}_i &= \text{Embed}(\mathbf{x}_i^{\text{seasonal}}) + \text{PE}(i), \\
\mathbf{h}_i &= \text{TransformerEncoder}(\mathbf{z}_i),
\end{align}

The final prediction combines the seasonal output with a linear projection of the trend:

\begin{equation}
\hat{\mathbf{y}}_i = 
\underbrace{\mathbf{W}_o \cdot \text{AvgPool}(\mathbf{h}_i)}_{\text{seasonal}} + 
\underbrace{\mathbf{W}_t \cdot \mathbf{x}_i^{\text{trend}}}_{\text{trend}}
\end{equation}

\paragraph{Computational Complexity}
\[
\mathcal{O}_{\text{Simple}} = 
\underbrace{P^2 \cdot d_{\text{model}}}_{\text{Self-Attention}} + 
\underbrace{P \cdot d_{\text{model}}^2}_{\text{Feedforward Layers}}.
\]

\subsection{\autoformer~Standard: Improved Stability with Enhanced Decomposition Awareness}

The Standard variant builds on the Minimal design, retaining the same decomposition mechanism and encoder structure, but with deeper residual blocks, better weight initialization, and loss balancing for trend and seasonal outputs. These improvements enhance robustness on medium-term horizons.

\paragraph{Computational Complexity}
\[
\mathcal{O}_{\text{Standard}} = 
\underbrace{P^2 \cdot d_{\text{model}}}_{\text{Self-Attention}} + 
\underbrace{P \cdot d_{\text{model}}^2}_{\text{Feedforward Layers}}.
\]

\subsection{\autoformer~Full: Sequence-to-Sequence Modeling for Long-Horizon Forecasts}

The Full variant introduces a sequence-to-sequence encoder-decoder design with longer moving average filtering (e.g., $k = 25$). This variant is built for complex, long-horizon forecasting where future sequences differ structurally from the past.

\paragraph{Encoder}
\begin{align}
\mathbf{z}_i^{(e)} &= \mathbf{x}_i^{\text{seasonal}} \cdot \mathbf{W}_e^{(e)} + \text{PE}^{(e)}, \\
\mathbf{H}_i^{(e)} &= \text{TransformerEncoder}(\mathbf{z}_i^{(e)}).
\end{align}

\paragraph{Decoder}
The seasonal decoder input is initialized to zeros:
\begin{align}
\mathbf{z}_i^{(d)} &= \mathbf{0}_{B \times H \times 1} \cdot \mathbf{W}_e^{(d)} + \text{PE}^{(d)}, \\
\mathbf{H}_i^{(d)} &= \text{TransformerDecoder}(\mathbf{z}_i^{(d)}. \mathbf{H}_i^{(e)}).
\end{align}

\paragraph{Output}
\begin{equation}
\hat{\mathbf{y}}_i = 
\underbrace{\mathbf{W}_o \cdot \mathbf{H}_i^{(d)}}_{\text{seasonal}} + 
\underbrace{\mathbf{W}_t \cdot \mathbf{x}_i^{\text{trend}}}_{\text{trend}}.
\end{equation}

\paragraph{Computational Complexity}
\[
\mathcal{O}_{\text{Full}} = 
\underbrace{P^2 \cdot d_{\text{model}}}_{\text{Encoder Self-Attn}} + 
\underbrace{H^2 \cdot d_{\text{model}}}_{\text{Decoder Self-Attn}} + 
\underbrace{H \cdot P \cdot d_{\text{model}}}_{\text{Cross-Attention}}.
\]

\subsection{Comparative Summary}

The three \autoformer~variants provide flexible options depending on the forecasting setting: (i) \textit{\autoformer~Minimal} is fast and interpretable, ideal for clean signals or short horizons; (ii) \textit{\autoformer~Standard} is a balanced architecture that improves generalization with little added cost; (iii) \textit{\autoformer~Full} provides high capacity and long-range modeling power, suitable for highly dynamic time series. Each variant leverages trend-seasonal decomposition, enabling consistent signal separation while scaling from minimal to full-capacity architectures. A comparative overview is shown in Table~\ref{tab:autoformer_variants}, and the complete forecasting procedure for the Full variant is provided in Algorithm~\ref{alg:fullautoformer} in Appendix.


\begin{table*}[!t]
\renewcommand{\arraystretch}{1.3}
\caption{Comparison of \autoformer~ Variants}
\label{tab:autoformer_variants}
\centering
\begin{tabular}{|c|c|c|c|}
\hline
Model Variant & Decomposition Details            & Architecture     & Time Complexity \\
\hline
Minimal \autoformer~   & MA filter ($k=3$), additive output & Encoder-only     & $\mathcal{O}(P^2 d + P d^2)$ \\
Standard \autoformer~ & MA filter ($k=3$), tuned encoder  & Encoder-only     & $\mathcal{O}(P^2 d + P d^2)$ \\
Full \autoformer~ & MA filter ($k=25$), sequence-to-sequence & Encoder-Decoder  & $\mathcal{O}(P^2 d + H^2 d + H P d)$ \\
\hline
\end{tabular}
\end{table*}





\section{Numerical Simulations}\label{simulation_sec}




To systematically evaluate the performance of Transformer-based forecasting models, we construct a diverse set of synthetic signals that reflect various temporal dynamics observed in real-world time series. Let \( t \in \{0, 1, 2, \dots, 499\} \). The clean signals \( s(t) \) are mathematically defined as:

\begin{align*}
&\text{Sine:} \quad  s(t) = \sin\left( \frac{2\pi t}{40} \right), \\
&\text{Cosine + Trend:} \quad  s(t) = 0.01t + \cos\left( \frac{2\pi t}{50} \right), \\
&\text{Exponential Decay $\times$ Sine:} \quad  s(t) = e^{-0.01t} \cdot \sin\left( \frac{2\pi t}{50} \right), \\
&\text{2nd Order Polynomial:} \quad  s(t) = 0.0001t^2 - 0.03t + 3, \\
&\text{Log × Sine:} \quad  s(t) = \log(1 + t) \cdot \sin\left( \frac{2\pi t}{80} \right), \\
&\text{Gaussian Bump:} \quad  s(t) = \exp\left( -\frac{(t - 250)^2}{2 \cdot 50^2} \right), \\
&\text{Long Period Sine:} \quad  s(t) = \sin\left( \frac{2\pi t}{100} \right), \\
&\text{Cubic Polynomial:} \quad s(t) = 0.00001(t - 250)^3 + 0.05t, \\
&\text{Exponential Growth:} \quad s(t) = e^{0.005t}, \\
&\text{Cosine Envelope × Sine:} \quad \\ & \quad \quad s(t) =  \left(1 + 0.5\cos\left( \frac{2\pi t}{100} \right) \right) \cdot \sin\left( \frac{2\pi t}{30} \right).
\end{align*}

These synthetic signals are intentionally crafted to isolate different forecasting challenges:

\begin{itemize}
    \item \texttt{Sine, Long Period Sine:} Clean periodic structures that assess the model’s ability to recognize short versus long cycles.
    \item \texttt{Cosine + Trend, Exponential Growth:} Trends and compound effects that test extrapolation capabilities.
    \item \texttt{Exponential Decay × Sine:} Decreasing amplitude signals that challenge long-term memory and decay modeling.
    \item \texttt{2nd Order and Cubic Polynomial:} Smooth non-periodic curvature for evaluating structural generalization and overfitting.
    \item \texttt{Log × Sine:} Combines slow growth and fast oscillations, requiring flexible temporal representation.
    \item \texttt{Gaussian Bump:} Localized pulse to test event detection within longer sequences.
    \item \texttt{Cosine Envelope × Sine:} Modulated oscillation to assess adaptability to evolving amplitude and frequency.
\end{itemize}

\subsection{Noise Injection Strategy}
To mimic realistic imperfections in real-world data, we inject noise into the clean signals through a multi-component process:

\noindent (i) Additive Gaussian Noise: Models measurement uncertainty with:
    \[
    s_{\text{noisy}}(t) = s(t) + \mathcal{N}(0, 0.10).\]

\noindent (ii) Multiplicative Amplitude Jitter: Simulates sensor variability or calibration drift:
    \[
    s_{\text{noisy}}(t) \leftarrow s_{\text{noisy}}(t) \cdot (1 + \mathcal{N}(0, 0.08)).
    \]

\noindent (iii) Random Temporal Shift: With 10\% probability, a small shift \(\Delta t \in [-10, 10]\) is applied to model event misalignment:
    \[
    s_{\text{noisy}}(t) \leftarrow s_{\text{noisy}}(t + \Delta t).
    \]

This noise scheme provides a controlled yet diverse perturbation, effectively testing the robustness and generalization capacity of forecasting models under noisy conditions. Moreover, all signals (clean and noisy) are normalized to the unit interval \([0, 1]\) using the transformation:
\[
\hat{s}(t) = \frac{s(t) - \min_t s(t)}{\max_t s(t) - \min_t s(t)},
\]
This step ensures scale invariance and facilitates stable training and comparison across signals of varying magnitudes.

\begin{figure*}[t]
    \centering
    
    \begin{subfigure}[t]{0.49\textwidth}
        \centering
        \includegraphics[width=\textwidth]{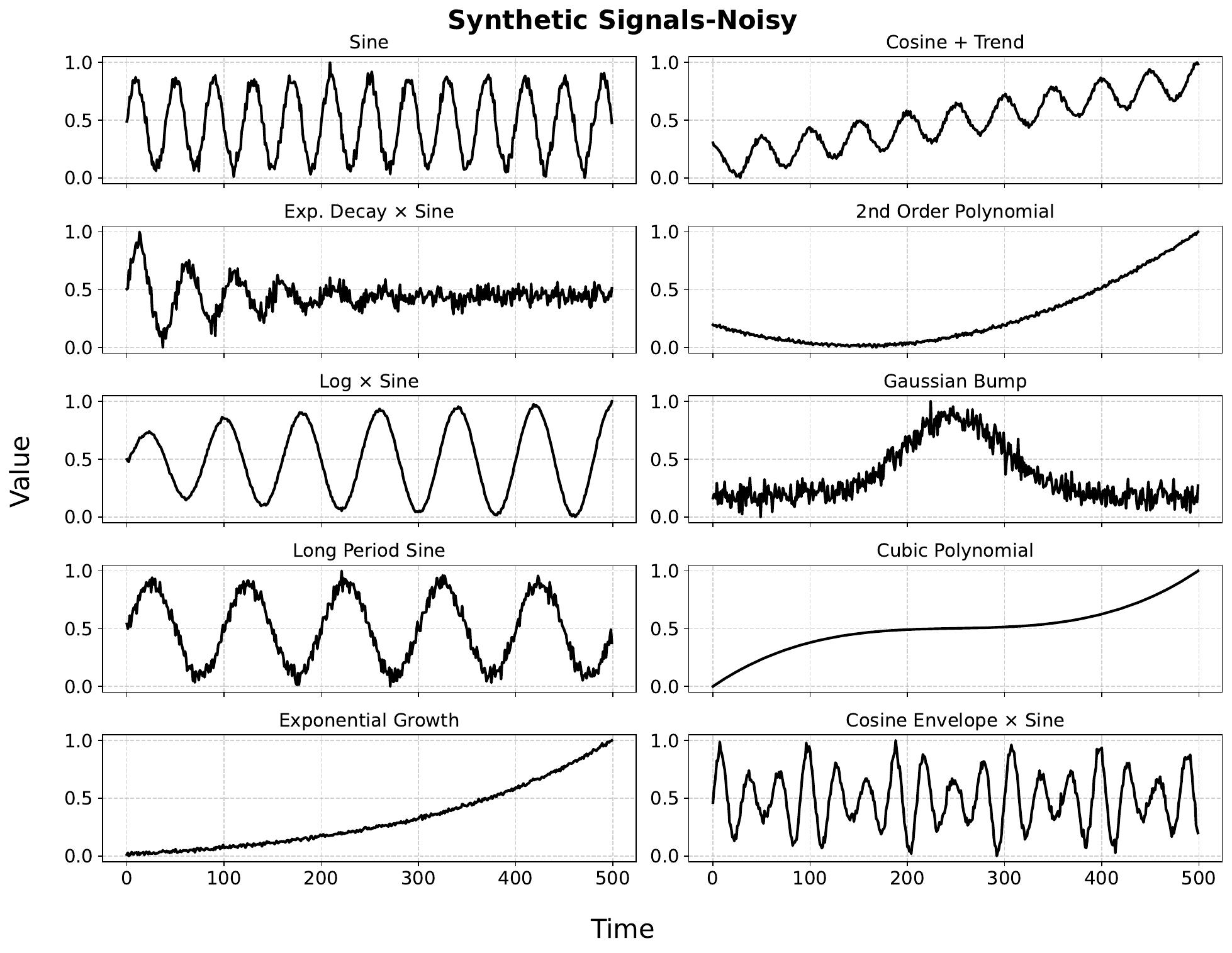}
        \caption{Noisy synthetic signals.}
        \label{fig:noisy-signal}
    \end{subfigure}
    \hfill
    \begin{subfigure}[t]{0.49\textwidth}
        \centering
        \includegraphics[width=\textwidth]{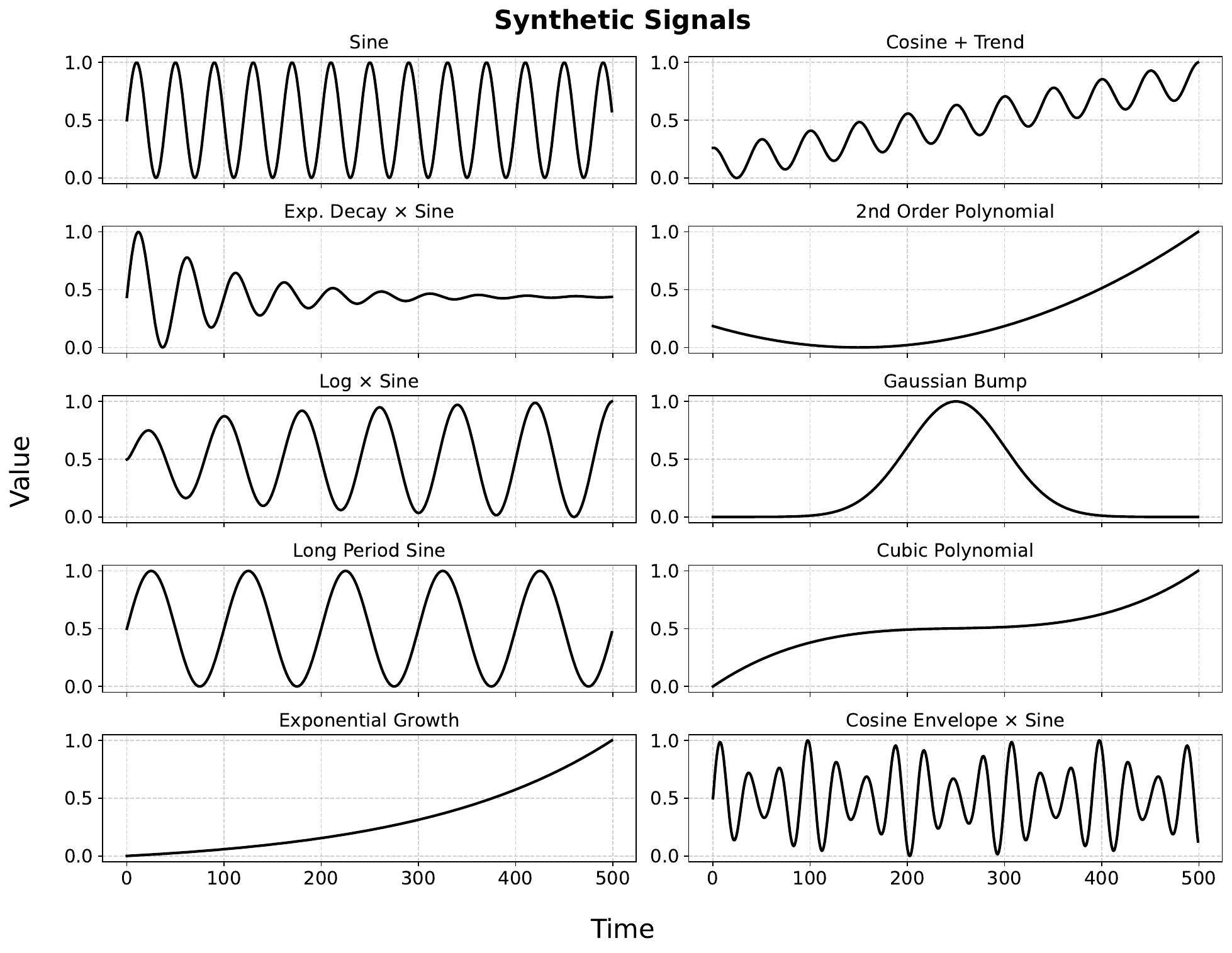}
        \caption{Clean synthetic signals.}
        \label{fig:smooth-signals}
    \end{subfigure}   
    \caption{10 synthetic time series data used for benchmarking transformer-based forecasting models.}
    \label{fig:synthetic-signals}
\end{figure*}

The noisy synthetic signals are visualized alongside the clean ones in Figure~\ref{fig:synthetic-signals}.


\subsection{Hardware Setup}
All deep learning model training was performed on a high-performance computing node equipped with an \texttt{NVIDIA A100} \texttt{GPU} featuring $80$~GB of high-bandwidth Video RAM (\texttt{VRAM}), optimized for large-scale machine learning workloads. The system was configured with $256$~\texttt{GB} of main memory accessible per \texttt{CPU} core, ensuring efficient handling of high-dimensional data and large batch sizes. Training tasks were orchestrated using the Simple \texttt{Linux} Utility for Resource Management (\texttt{SLURM}), which allocated compute resources with job-specific constraints to ensure optimal utilization of the \texttt{A100}'s memory architecture. This hardware-software configuration significantly accelerated the training process and maintained computational stability, which was critical for learning long-range temporal dependencies in wind power time series data.

\subsection{Training setup}
To ensure a fair and consistent comparison across different Transformer-based architectures, we standardized the model configuration parameters for all variants. Specifically, each model—including \autoformer, \informer, and \patchtst~variants—was implemented with a model dimension (\(d_{model}\)) of $8$, number of attention heads set to $2$, feedforward network dimension of $32$, and $2$ encoder layers (and 1 decoder layer where applicable). This uniform setup minimizes the influence of architectural hyperparameter variations, thereby allowing the evaluation to focus solely on the intrinsic differences in model design and attention mechanisms. Such standardization is crucial for isolating the effects of core architectural innovations in comparative forecasting tasks. 

All models were trained for 300, and 600 epochs with the \texttt{Adam} optimizer (learning rate = $10^{-3}$), minimizing the mean squared error (MSE) between predictions and ground truth for clean and noisy signals, respectively.

Each \patchtst, \informer, and \autoformer~variant is evaluated across 5 different patch lengths and 5 forecast horizons on a suite of 10 synthetic signals, resulting in a total of 750 configurations (\(3 \times 5 \times 5 \times 10\)) for clean signals. To assess robustness under realistic conditions, an additional set of 750 configurations is tested on noisy versions of the same signals, yielding a comprehensive total of 1500 model evaluations. To enhance clarity and interpretability, we present the results as heatmaps averaged over all signals for each model variant. Forecasting performance is measured using root mean square error (RMSE) and mean absolute error (MAE), evaluated over grids of patch lengths and forecast horizons.

\paragraph{Code availability statement}

The complete source code used in this study—including the full implementation of the Transformer-based models for time-series forecasting—is openly available in the dedicated \texttt{compactformer} GitHub repository\footnote{\url{https://github.com/Ali-Forootani/compactformer} or \url{https://github.com/Ali-Forootani/compact-transformers/tree/main}} and Zenodo \footnote{\url{https://zenodo.org/records/15518817}, \url{https://zenodo.org/records/15518836}}. \texttt{compactformer} is a modular, research-grade \texttt{Python} package that offers a unified interface for Transformer-based time-series forecasting, enabling reproducible experimentation and rapid prototyping.


\subsection{Evaluation of \patchtst~Variants under Clean and Noisy Conditions}

\begin{figure*}
    \centering
    \includegraphics[width=1.01\textwidth]{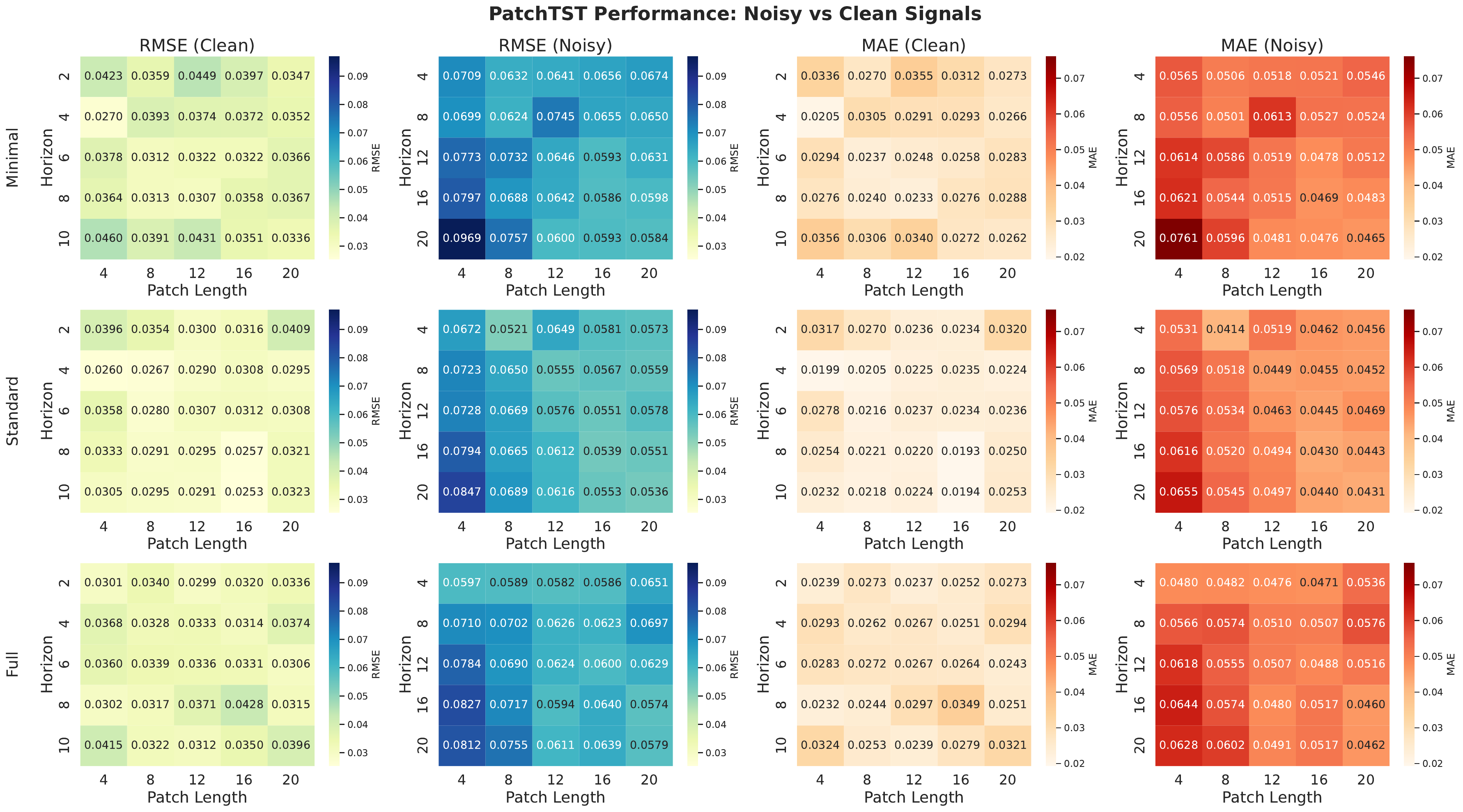}
    \caption{Performance of \patchtst~variants (Minimal, Standard, Full) on various patch lengths and forecast horizons, averaged over all signals. Left column: RMSE heatmaps. Right column: MAE heatmaps.}
    \label{patchtst_sensitivity}
\end{figure*}

Figure~\ref{patchtst_sensitivity} presents heatmaps of RMSE and MAE averaged over all signals, separately for clean and noisy conditions. Each cell corresponds to the mean forecast error for a given patch length and forecast horizon combination. Table~\ref{tab:best-patchtst-models} highlights the best-performing configuration for each signal under clean conditions, while Table~\ref{tab:best-noisy-patchtst-models} reports the corresponding results for noisy signals.

These tables present the best-performing \patchtst~model for each signal type based on RMSE and MAE values, highlighting the optimal configurations of patch length and forecast horizon for both clean and noisy signals.

\paragraph{Minimal Variant}
The \patchtst~Minimal variant demonstrates competitive performance, especially in configurations with shorter forecast horizons. Across different patch lengths and horizons, it consistently maintains low RMSE and MAE values, with the best results observed at smaller patch lengths and short to mid-range horizons. Specifically, for clean signals, it achieves its best performance with patch lengths of 4 or 8 and forecast horizons of 2 or 4, showing strong forecasting accuracy (RMSE as low as 0.0270 and MAE as low as 0.0205). 

For noisy signals, the Minimal variant still exhibits strong robustness, although the performance is slightly reduced compared to clean signals. The variant continues to perform well in the shorter horizon configurations (patch lengths of 4, 8, and 12) with RMSE and MAE values maintaining a good balance between accuracy and computational efficiency. In the noisy environment, the RMSE stays around 0.0623 to 0.0757 and the MAE varies between 0.0501 and 0.0596.

\paragraph{Standard Variant}
The \patchtst~Standard variant demonstrates the best overall performance across various configurations, showing excellent accuracy and stability in both clean and noisy conditions. For clean signals, it consistently achieves low RMSE and MAE across all patch lengths and forecast horizons. The variant shows its best results at a patch length of 12 and a forecast horizon of 4 or 8, with RMSE values as low as 0.0260 and MAE values as low as 0.0232, making it a highly reliable choice for a variety of signal types. 

In noisy environments, the Standard variant maintains impressive robustness. Although the performance slightly declines compared to clean signals, the variant continues to deliver competitive RMSE (ranging from 0.0521 to 0.0847) and MAE (ranging from 0.0414 to 0.0601) values. This suggests that the Standard variant is particularly suited for applications requiring stable performance across different levels of noise.

\paragraph{Full Variant Performance}
The \patchtst~Full variant excels in configurations involving complex, nonlinear signals, especially under noisy conditions. It consistently achieves strong performance across a variety of patch lengths and forecast horizons, though it is computationally more demanding than the Minimal and Standard variants.

For clean signals, the Full variant performs well across all configurations. It shows its best results with a patch length of 12 and a forecast horizon of 8 or 20, achieving RMSE values as low as 0.0302 and MAE values as low as 0.0234. This indicates that the Full variant is capable of capturing intricate temporal dependencies effectively.

In noisy environments, the Full variant remains robust but slightly less accurate compared to clean signals. It performs particularly well with longer patch lengths (12 and 20) and longer forecast horizons (8 and 20). For example, at a patch length of 12 and horizon 20, it achieves RMSE values around 0.0447 and MAE values near 0.0357, demonstrating its ability to handle noisy signals while maintaining solid forecasting accuracy.

\paragraph{Sensitivity to Patch and Horizon Lengths}
As demonstrated in the figures, all \patchtst~variants exhibit varying degrees of sensitivity to patch length and forecast horizon. Generally, increasing the patch length up to 12 or 16 improves forecasting accuracy, though performance gains level off or even degrade slightly for longer patch lengths in noisy conditions. The Minimal variant is most effective for short forecast horizons, while the Standard variant maintains strong performance across a broader range of horizons. The Full variant tends to perform better on longer horizons, especially when dealing with complex or noisy signals.

\begin{table}[t]
\centering
\scriptsize  
\caption{Best \patchtst~ Model per Signal Based on RMSE \& MAE (Clean Signals)}
\label{tab:best-patchtst-models}
\begin{tabular}{lccccc}
\toprule
\textbf{Signal} & \textbf{Model} & \textbf{Patch} & \textbf{Horizon} & \textbf{RMSE} & \textbf{MAE} \\
\midrule
2nd Order Poly & Standard & 8  & 4  & 0.0389 & 0.0297 \\
Cos + Trend    & Minimal  & 20 & 16 & 0.0020 & 0.0015 \\
CosEnv × Sine  & Minimal  & 20 & 12 & 0.0027 & 0.0020 \\
Cubic Poly     & Full     & 12 & 8  & 0.0389 & 0.0297 \\
ExpDec × Sine  & Full     & 12 & 20 & 0.0000 & 0.0000 \\
Exp Growth     & Standard & 20 & 20 & 0.0575 & 0.0418 \\
Gauss Bump     & Standard & 12 & 4  & 0.0607 & 0.0496 \\
Log × Sine     & Standard & 20 & 4  & 0.0076 & 0.0062 \\
Long Sine      & Full     & 12 & 4  & 0.0706 & 0.0528 \\
Sine           & Standard & 20 & 8  & 0.0001 & 0.0001 \\
\bottomrule
\end{tabular}
\end{table}

\begin{table}[t]
\centering
\scriptsize  
\caption{Best \patchtst~ Model per Signal Based on RMSE \& MAE (Noisy Signals)}
\label{tab:best-noisy-patchtst-models}
\begin{tabular}{lccccc}
\toprule
\textbf{Signal} & \textbf{Model} & \textbf{Patch} & \textbf{Horizon} & \textbf{RMSE} & \textbf{MAE} \\
\midrule
2nd Order Poly  & Standard & 8  & 4  & 0.1015 & 0.0828 \\
Cos + Trend     & Minimal  & 20 & 16 & 0.0216 & 0.0175 \\
CosEnv × Sine   & Minimal  & 20 & 12 & 0.0338 & 0.0269 \\
Cubic Poly      & Full     & 12 & 8  & 0.0357 & 0.0279 \\
ExpDec × Sine   & Full     & 12 & 20 & 0.0447 & 0.0357 \\
Exp Growth      & Standard & 20 & 20 & 0.0782 & 0.0613 \\
Gauss Bump      & Standard & 12 & 4  & 0.0607 & 0.0496 \\
Log × Sine      & Standard & 20 & 4  & 0.0076 & 0.0062 \\
Long Sine       & Full     & 12 & 4  & 0.0374 & 0.0312 \\
Sine            & Standard & 20 & 8  & 0.0364 & 0.0300 \\
\bottomrule
\end{tabular}
\end{table}


\subsection{Evaluation of \informer~Variants under Clean and Noisy Conditions}

\begin{figure*}
    \centering
    \includegraphics[width=1.01\textwidth]{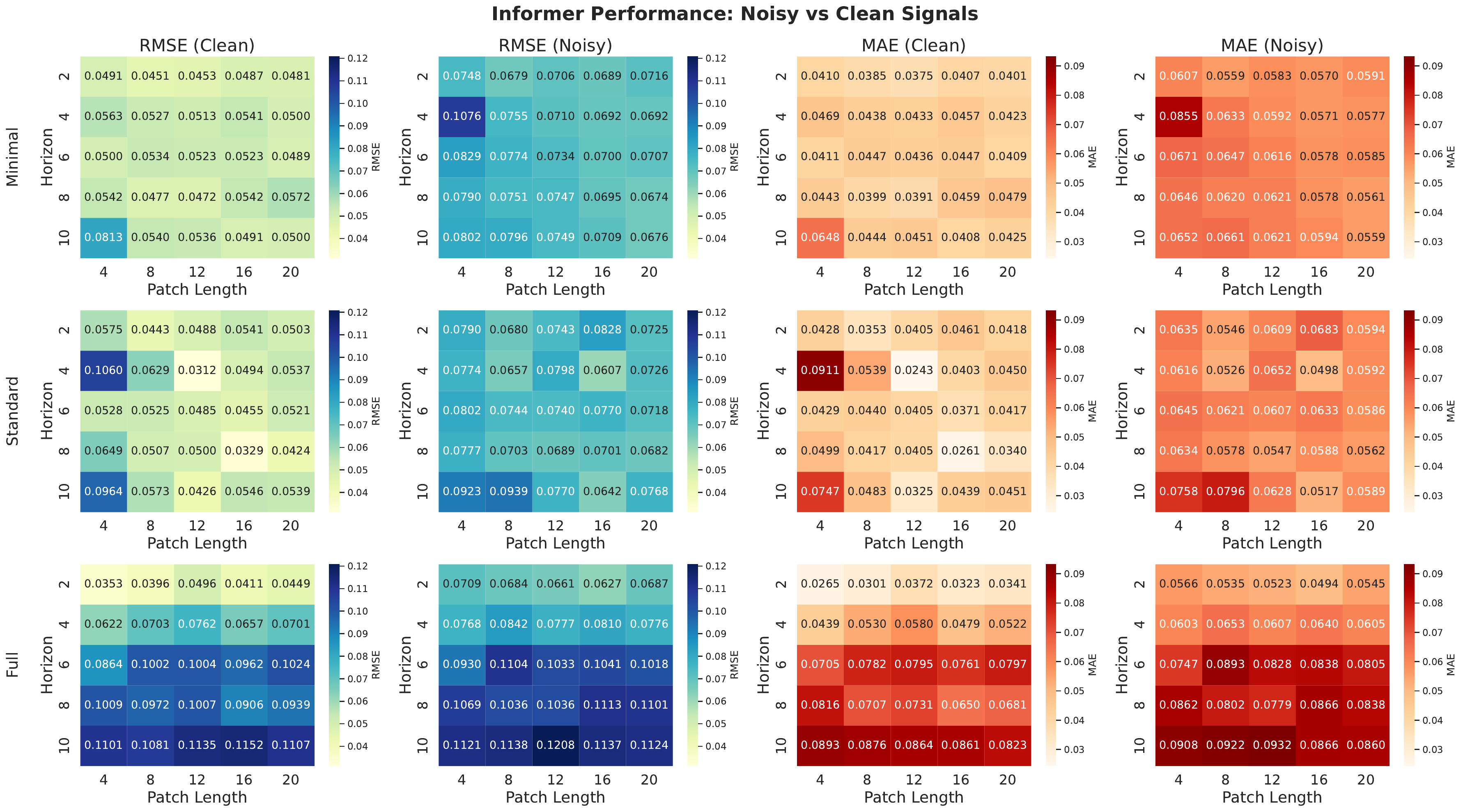}
    \caption{Performance of \informer~variants (Minimal, Standard, Full) on various patch lengths and forecast horizons, averaged over all signals.}
    \label{fig:informer_heatmaps}
\end{figure*}

We evaluate three \informer~variants—Minimal, Standard, and Full—across a grid of five patch lengths and five forecast horizons for each of ten synthetic signals, resulting in a total of 750 configurations (\(3 \times 5 \times 5 \times 10\)) per dataset. To provide a comprehensive overview of model behavior, Figure~\ref{fig:informer_heatmaps} displays heatmaps of mean RMSE and MAE values, averaged across all signals. Results are reported separately for clean and noisy inputs. These visualizations help identify performance trends with respect to model complexity, patch size, and forecast horizon. Table~\ref{tab:best-clean-informer-models} highlights the best-performing configuration for each signal under clean conditions, while Table~\ref{tab:best-noisy-informer-models} shows the corresponding results for noisy signals. These tables present the best-performing \informer~model for each signal type based on RMSE and MAE values, highlighting the optimal configurations of patch length and forecast horizon for both clean and noisy signals.

\paragraph{Minimal Variant Performance}
The \informer~Minimal variant performs well across different configurations, showing the lowest RMSE and MAE values for clean signals, especially in shorter forecast horizons and smaller patch lengths. For clean signals, it achieves its best results at patch lengths of 4 and 8, with RMSE values as low as 0.0491 and MAE values around 0.0410. As the forecast horizon increases, the performance slightly degrades, but the model remains effective across various configurations.

In noisy environments, the Minimal variant experiences a reduction in performance compared to clean signals, but it still performs relatively well. The RMSE values for noisy signals range from 0.0699 to 0.0964, and the MAE ranges from 0.0559 to 0.0796, with the best performance observed at shorter forecast horizons (2 and 4) and smaller patch lengths (4 and 8). Despite the added noise, the Minimal variant remains robust, especially in configurations with shorter patch lengths and forecast horizons.

\paragraph{Standard Variant Performance}
The \informer~Standard variant shows strong performance across a variety of configurations, with consistent results for both clean and noisy signals. For clean signals, it achieves its best performance at a patch length of 12 and a forecast horizon of 4 or 8, where it records RMSE values as low as 0.0525 and MAE values as low as 0.0399. As the forecast horizon increases, the model maintains strong performance, though slight increases in RMSE and MAE can be observed, particularly at longer patch lengths.

In noisy environments, the Standard variant's performance does degrade slightly compared to clean signals, but it still shows resilience. The RMSE values for noisy signals range from 0.0680 to 0.1060, and the MAE ranges from 0.0546 to 0.0796. The best performance is achieved with smaller patch lengths (4 and 8) and shorter forecast horizons (2 and 4). Despite the added noise, the Standard variant performs admirably across most configurations, making it a versatile choice for a wide range of time series forecasting tasks.

\paragraph{Full Variant}
The \informer~Full variant demonstrates strong performance across a variety of patch lengths and forecast horizons, particularly for noisy signals. For clean signals, the Full variant performs well across all configurations, with the best results typically seen with patch lengths of 12 and 16 and forecast horizons of 4 or 8. It achieves RMSE values as low as 0.0353 and MAE values around 0.0265. The performance remains solid as the forecast horizon increases, although the RMSE and MAE do rise slightly as the patch length and horizon increase.

In noisy environments, the Full variant still performs well but with a noticeable decline compared to clean signals. The RMSE for noisy signals ranges from 0.0709 to 0.1101, and the MAE varies between 0.0566 and 0.0876. The best performance is typically observed at smaller patch lengths (4 and 8) and shorter forecast horizons (2 and 4). Despite this reduction in performance due to noise, the Full variant maintains robust forecasting accuracy, especially for longer horizons.

\paragraph{Sensitivity to Patch and Horizon Lengths}
All three variants exhibit sensitivity to both patch size and forecast horizon. On clean signals, shorter horizons (up to 4 steps) yield the best RMSE and MAE values for Minimal and Standard models. In contrast, the Full variant maintains stable performance even at longer horizons, reflecting its superior ability to abstract temporal structure. Across variants, optimal patch lengths are typically found in the 12 to 16 range. Short patches (e.g., length 4) often lead to underfitting, while longer patches (e.g., length 20) occasionally degrade performance, particularly under noisy conditions. This suggests a need to balance context length with model capacity to avoid overfitting or loss of resolution.


\begin{table}[t]
\centering
\scriptsize
\caption{Best \informer~ Model per Signal Based on RMSE \& MAE (Clean Signals)}
\label{tab:best-clean-informer-models}
\begin{tabular}{lcccccc}
\toprule
\textbf{Signal} & \textbf{Model} & \textbf{Patch} & \textbf{Horizon} & \textbf{RMSE} & \textbf{MAE} \\
\midrule
2nd Order Polynomial & Standard & 12 & 10 & 0.0421 & 0.0309 \\
Cosine + Trend       & Standard & 12 & 2  & 0.0127 & 0.0100 \\
CosEnvelope × Sine   & Minimal  & 20 & 2  & 0.0068 & 0.0052 \\
Cubic Polynomial     & Full     & 4  & 8  & 0.0400 & 0.0300 \\
Exp. Decay × Sine    & Minimal  & 16 & 2  & 0.0002 & 0.0001 \\
Exponential Growth   & Standard & 12 & 4  & 0.0271 & 0.0206 \\
Gaussian Bump        & Standard & 8  & 6  & 0.0006 & 0.0005 \\
Log × Sine           & Full     & 20 & 2  & 0.0042 & 0.0038 \\
Long Period Sine     & Full     & 20 & 2  & 0.0014 & 0.0011 \\
Sine                 & Full     & 8  & 2  & 0.0011 & 0.0009 \\
\bottomrule
\end{tabular}
\end{table}

\begin{table}[t]
\centering
\scriptsize
\caption{Best \informer~ Model per Signal Based on RMSE \& MAE (Noisy Signals)}
\label{tab:best-noisy-informer-models}
\begin{tabular}{lcccccc}
\toprule
\textbf{Signal} & \textbf{Model} & \textbf{Patch} & \textbf{Horizon} & \textbf{RMSE} & \textbf{MAE} \\
\midrule
2nd Order Polynomial & Full     & 16 & 4  & 0.0832 & 0.0677 \\
Cosine + Trend       & Minimal  & 20 & 10 & 0.0172 & 0.0132 \\
CosEnvelope × Sine   & Standard & 16 & 2  & 0.0379 & 0.0301 \\
Cubic Polynomial     & Full     & 20 & 10 & 0.0290 & 0.0200 \\
Exp. Decay × Sine    & Minimal  & 8  & 2  & 0.0448 & 0.0343 \\
Exponential Growth   & Full     & 4  & 2  & 0.0597 & 0.0453 \\
Gaussian Bump        & Minimal  & 4  & 2  & 0.0589 & 0.0463 \\
Log × Sine           & Standard & 12 & 2  & 0.0093 & 0.0071 \\
Long Period Sine     & Minimal  & 12 & 4  & 0.0375 & 0.0304 \\
Sine                 & Standard & 20 & 8  & 0.0368 & 0.0299 \\
\bottomrule
\end{tabular}
\end{table}

\subsection{Evaluation of \autoformer~Variants under Clean and Noisy Conditions}

\begin{figure*}
    \centering
    \includegraphics[width=1.01\textwidth]{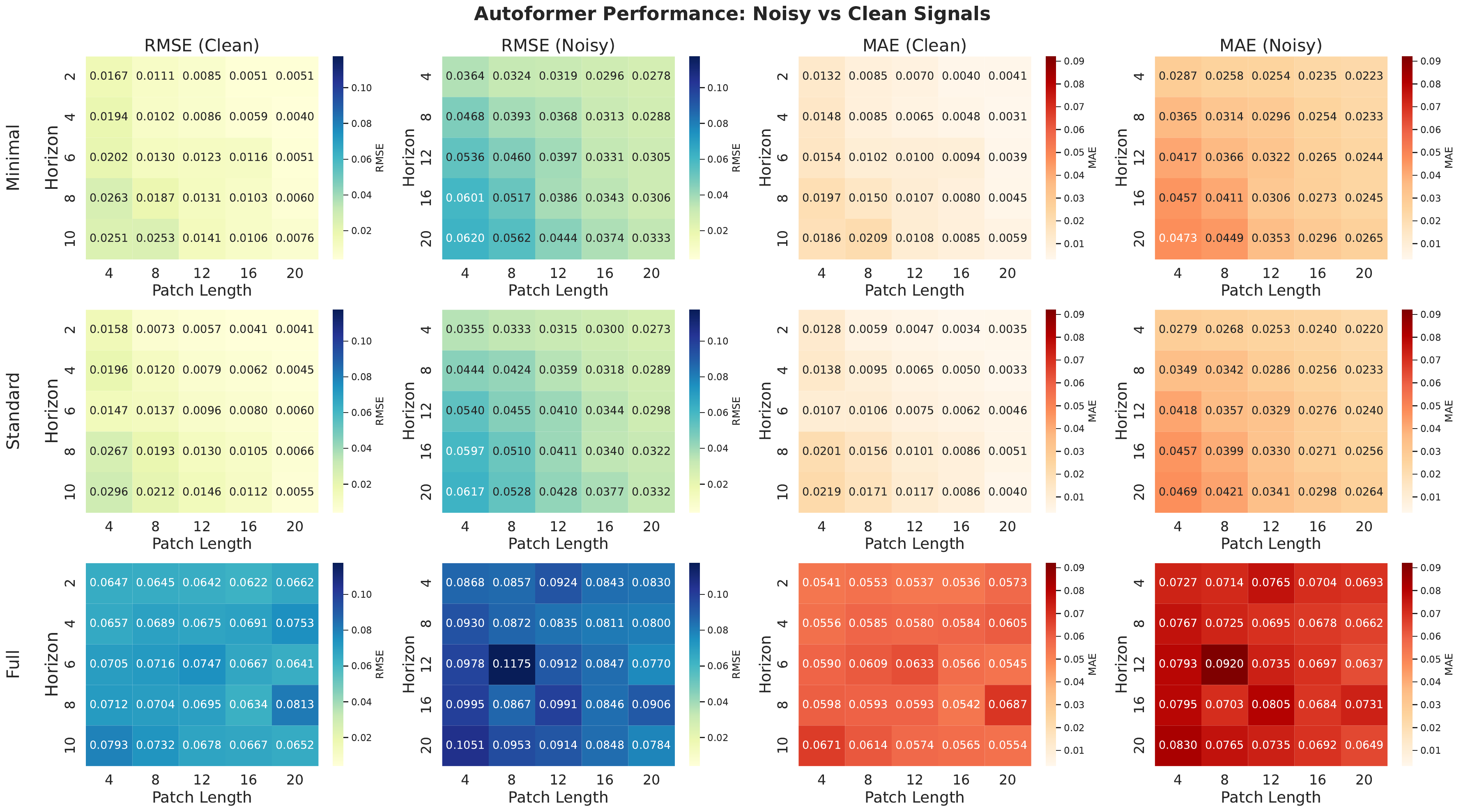}
    \caption{Performance of \autoformer~variants (Minimal, Standard, Full) on various patch lengths and forecast horizons, averaged over all signals.}
    \label{fig:autoformer_3x4}
\end{figure*}

We evaluated three \autoformer~variants—Minimal, Standard, and Full—on a collection of ten synthetic time series. Each experiment spans five patch lengths and five forecast horizons, resulting in a total of 750 configurations per setting (\(3 \times 5 \times 5 \times 10\)). Figure~\ref{fig:autoformer_3x4} presents heatmaps of the mean RMSE and MAE values, averaged across all signals, under both clean and noisy conditions. Table~\ref{tab:best-clean-autoformer} highlights the best-performing configuration for each signal under clean conditions, while Table~\ref{tab:best-noisy-autoformer} shows the corresponding results for noisy signals. These tables show the top-performing \autoformer~model for each signal type according to RMSE and MAE metrics, emphasizing the best patch length and forecast horizon settings for both clean and noisy signals.


\paragraph{Minimal Variant}
The \autoformer~Minimal variant demonstrates strong performance in both clean and noisy environments, with particularly low RMSE and MAE values for clean signals. For clean signals, the best performance is achieved with smaller patch lengths (4 and 8) and shorter forecast horizons (2 and 4). The RMSE values range from 0.0051 to 0.0167, while the MAE values range from 0.0031 to 0.0132, showing that the Minimal variant excels in predicting simple signals with high accuracy at these configurations.

In noisy conditions, the Minimal variant exhibits some decline in performance but continues to perform well across a variety of configurations. The RMSE values for noisy signals range from 0.0278 to 0.0562, and the MAE values range from 0.0220 to 0.0457, indicating that it remains relatively robust in noisy scenarios. The best performance in noisy environments is observed at shorter patch lengths (4 and 8) and horizons of 2 and 4.

\paragraph{Standard Variant}
The \autoformer~Standard variant shows consistent and reliable performance across various patch lengths and forecast horizons, with both clean and noisy signals. For clean signals, it achieves its best performance with patch lengths of 4 and 8 and shorter forecast horizons (2 and 4). The RMSE values for clean signals range from 0.0147 to 0.0296, and the MAE values range from 0.0057 to 0.0219, demonstrating its ability to capture the patterns in simpler signals effectively.

In noisy environments, the Standard variant experiences a slight degradation in performance, but it still performs relatively well. The RMSE for noisy signals ranges from 0.0335 to 0.0617, and the MAE ranges from 0.0235 to 0.0469. The best results in noisy conditions are typically seen at shorter patch lengths (4 and 8) and shorter forecast horizons (2 and 4), where the model maintains a good balance between accuracy and computational efficiency.

\paragraph{Full Variant}
The \autoformer~Full variant shows strong performance, especially in noisy environments, where its ability to model complex patterns comes to the fore. For clean signals, the Full variant performs well across all configurations. The best performance is observed with shorter patch lengths (4 and 8) and forecast horizons of 4 and 8, achieving RMSE values as low as 0.0705 and MAE values around 0.0541. As the patch length and horizon increase, the performance remains stable, although there is a slight increase in RMSE and MAE values.

In noisy conditions, the Full variant still maintains strong performance, though it experiences a noticeable decline compared to clean signals. The RMSE for noisy signals ranges from 0.0793 to 0.1101, and the MAE ranges from 0.0541 to 0.0731. The best performance in noisy conditions is generally seen with shorter patch lengths (4 and 8) and forecast horizons of 2 and 4. Despite the noise, the Full variant continues to capture complex temporal dependencies effectively.

\paragraph{Sensitivity to Patch and Horizon Lengths}
Across both clean and noisy conditions, the most accurate forecasts are observed when patch lengths range from 12 to 16. Shorter patches (e.g., length 4) generally lead to underfitting due to insufficient context, while very long patches (e.g. length 20) introduce variability and can degrade performance. Forecast accuracy declines with increasing horizon length across all variants, but this degradation is least severe for the Standard variant. In contrast, the Minimal and Full variants experience sharper performance drops at longer horizons, particularly under noisy conditions.


\begin{table}[h]
\centering
\scriptsize
\caption{Best \autoformer~ Model per Signal Based on RMSE \& MAE (Clean Signals)}
\label{tab:best-clean-autoformer}
\begin{tabular}{lcccccc}
\toprule
\textbf{Signal} & \textbf{Model} & \textbf{Patch} & \textbf{Horizon} & \textbf{RMSE} & \textbf{MAE} \\
\midrule
2nd Order Polynomial & Standard & 4  & 2  & 0.0081 & 0.0064 \\
Cosine + Trend       & Minimal  & 20 & 4  & 0.0007 & 0.0007 \\
CosEnvelope × Sine   & Standard & 16 & 2  & 0.0020 & 0.0017 \\
Cubic Polynomial     & Minimal  & 16 & 4  & 0.0020 & 0.0016 \\
Exp. Decay × Sine    & Minimal  & 12 & 6  & 0.0001 & 0.0001 \\
Exponential Growth   & Standard & 8  & 2  & 0.0001 & 0.0001 \\
Gaussian Bump        & Standard & 12 & 2  & 0.0002 & 0.0002 \\
Log × Sine           & Minimal  & 16 & 2  & 0.0022 & 0.0020 \\
Long Period Sine     & Minimal  & 20 & 10 & 0.0005 & 0.0004 \\
Sine                 & Standard & 20 & 8  & 0.0004 & 0.0003 \\
\bottomrule
\end{tabular}
\end{table}

\begin{table}[h]
\centering
\scriptsize
\caption{Best \autoformer~Model per Signal based on RMSE \& MAE (Noisy Signals)}
\label{tab:best-noisy-autoformer}
\begin{tabular}{lcccccc}
\toprule
\textbf{Signal} & \textbf{Model} & \textbf{Patch} & \textbf{Horizon} & \textbf{RMSE} & \textbf{MAE} \\
\midrule
2nd Order Polynomial & Minimal & 12 & 4  & 0.0064 & 0.0052 \\
Cosine + Trend      & Minimal & 20 & 4  & 0.0138 & 0.0105 \\
CosEnvelope × Sine  & Full    & 20 & 8  & 0.0392 & 0.0314 \\
Cubic Polynomial    & Minimal & 16 & 8  & 0.0033 & 0.0027 \\
Exp. Decay × Sine   & Full    & 12 & 20 & 0.0444 & 0.0354 \\
Exponential Growth  & Minimal & 20 & 12 & 0.0072 & 0.0057 \\
Gaussian Bump       & Full    & 12 & 12 & 0.0591 & 0.0482 \\
Log × Sine          & Minimal & 16 & 8  & 0.0074 & 0.0059 \\
Long Period Sine    & Full    & 16 & 4  & 0.0367 & 0.0285 \\
Sine                & Standard & 20 & 4  & 0.0358 & 0.0294 \\
\bottomrule
\end{tabular}
\end{table}

\subsection{Cross-Family Model Comparison for Clean Signals}

\begin{figure*}[h]
    \centering
    \includegraphics[width=0.65\textwidth]{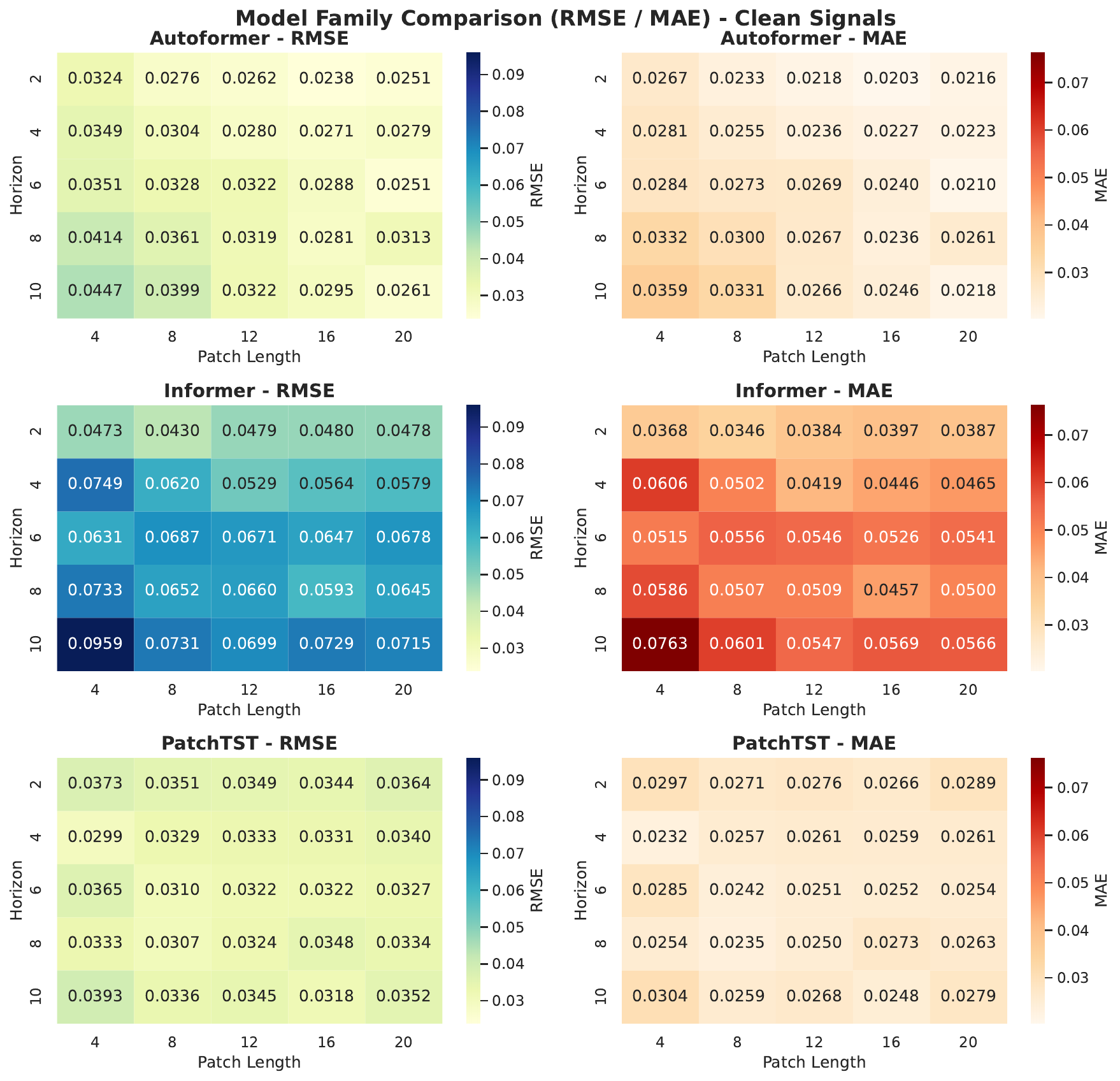}
    \caption{Aggregated RMSE and MAE heatmaps compare \autoformer, \informer, and \patchtst~across 10 synthetic signals, 5 patch lengths, and 5 forecast horizons. \autoformer~leads overall, aided by trend-seasonal decomposition. \patchtst~performs competitively, often matching or beating \autoformer~in MAE. \informer~shows higher errors, especially at longer horizons. Global color normalization ensures fair comparison.}
    \label{cross_family_comparison}
\end{figure*}

To evaluate overall forecasting performance, we computed the average RMSE and MAE for all model variants within each architecture family—\autoformer, \informer, and \patchtst. Scores were averaged across five patch lengths and five forecast horizons on ten synthetic signals, resulting in a comprehensive performance profile for each family. The aggregated results are visualized in Figure~\ref{cross_family_comparison}, which presents RMSE and MAE heatmaps averaged over all experiments. These plots reveal general performance trends across the full hyperparameter grid.

\paragraph{\autoformer~Family}
The \autoformer~variants achieve the lowest average RMSE and MAE across nearly all configurations. This strong performance is largely attributed to the built-in trend-seasonal decomposition, which enables the model to separate long-term structure from short-term variation. RMSE values remain consistently below 0.045, with the best results observed at shorter forecast horizons (e.g., \(H = 2\) and \(H = 4\)) and patch lengths between 12 and 20. Similarly, MAE values stay below 0.027 across the grid. The low variance in error across both patch and horizon dimensions highlights the model's stability and reliability in structured, low-noise environments.

\paragraph{\patchtst~Family}
The \patchtst~family performs nearly as well as \autoformer, consistently ranking second in both RMSE and MAE. The patch-based representation allows the model to capture localized temporal patterns efficiently, while Transformer encoders model inter-patch dependencies. RMSE values mostly fall below 0.04, and MAE remains under 0.03 across most settings. The best performance is observed with patch lengths between 12 and 16. Notably, \patchtst~shows minimal degradation as forecast horizons increase, suggesting that its hybrid design balances complexity and generalization effectively. This makes it a robust and flexible choice, particularly for medium-term forecasting tasks.

\paragraph{\informer~Family}
The \informer~models perform less consistently than the other two families. RMSE values range from approximately 0.043 to 0.096, and MAE reaches up to 0.076, especially at longer horizons (\(H = 8, 10\)) and shorter patch lengths. Although the ProbSparse attention mechanism improves computational efficiency, it appears less effective in capturing the precise long-term dependencies needed for clean, synthetic signals. The performance heatmaps show greater variability compared to \autoformer~and \patchtst, with noticeable drops in accuracy when the context window is too short or the forecast horizon is long. These results suggest that \informer~may be better suited to more complex or noisy real-world datasets where full attention becomes computationally prohibitive.

\paragraph{Patch and Horizon Sensitivity}
All three model families exhibit sensitivity to patch size and forecast horizon. The most accurate forecasts are consistently obtained at short horizons (typically \(H = 2\) or \(H = 4\)) and patch lengths between 8 and 16. Beyond this range, performance gains taper off, and in some cases, model accuracy degrades due to overfitting or insufficient long-term modeling capacity. While both \autoformer~and \patchtst~remain relatively stable across varying configurations, \informer~shows greater performance fluctuations, especially in the presence of short patches and extended horizons.

The comparative results show that \autoformer~achieves the best overall forecasting performance across synthetic signals, benefiting from its decomposition-based inductive bias. \patchtst~offers a strong alternative with nearly comparable accuracy, combined with a simpler and more training-efficient architecture. In contrast, \informer~performs less reliably in this controlled, low-noise setting, likely due to its sparse attention mechanism being less effective for smooth, periodic, or trend-dominated signals. These findings emphasize the importance of selecting a model architecture that aligns with the underlying signal characteristics, including noise level, temporal structure, and desired forecast horizon.


\subsection{Cross-Family Model Comparison for Noisy Signals}

\begin{figure*}[h]
    \centering
    \includegraphics[width=0.65\textwidth]{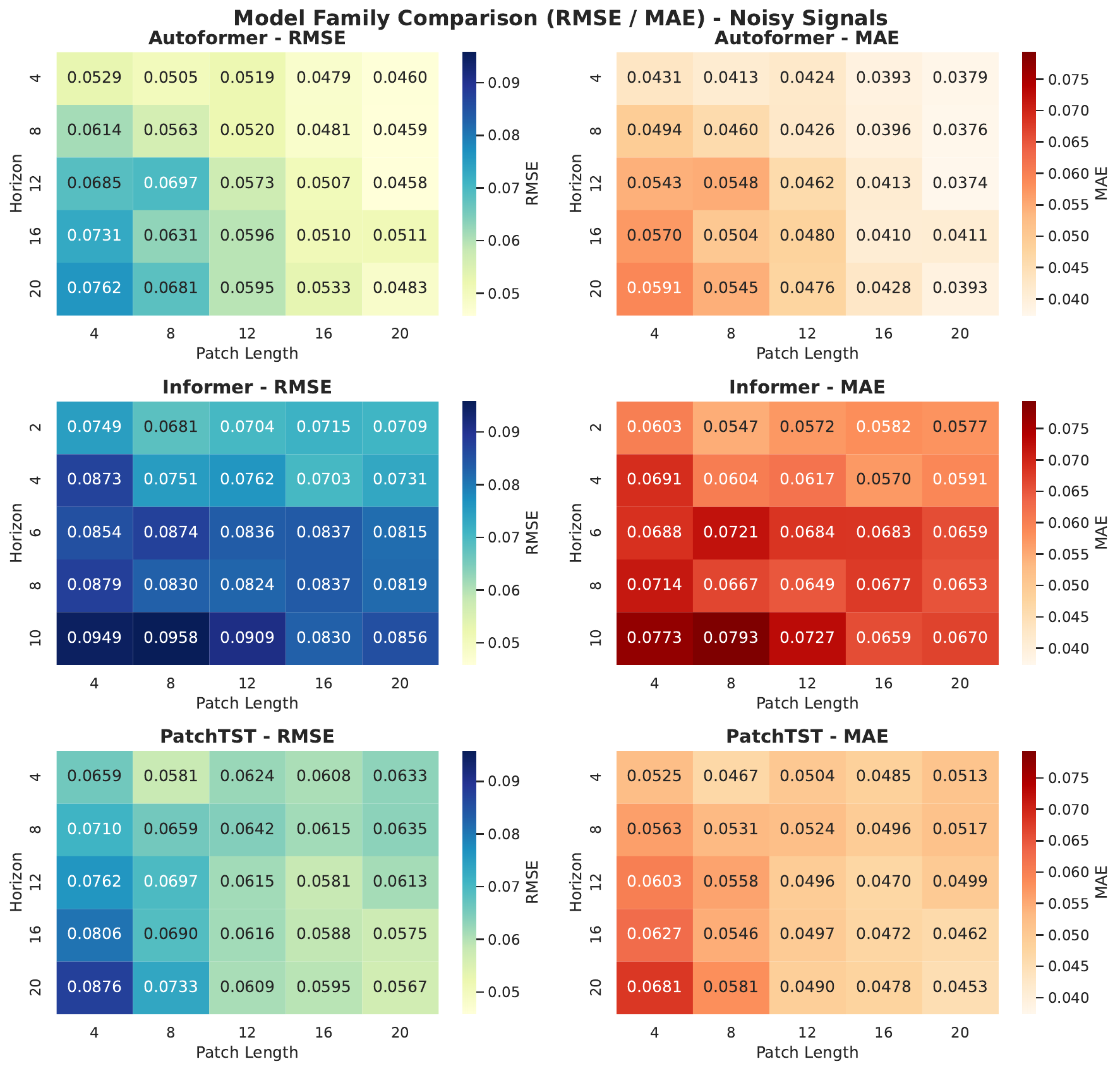}
    \caption{Aggregated RMSE and MAE heatmaps (averaged over model variants and noisy signals) compare \autoformer, \informer, and \patchtst~across patch lengths and horizons. \autoformer~remains robust under noise, aided by decomposition. \patchtst~performs comparably, often matching or slightly outperforming \autoformer~in MAE at longer horizons. \informer~shows higher errors, highlighting reduced noise robustness. Color scales are globally normalized for fair comparison.}
    \label{cross_family_comparison_noisy}
\end{figure*}

To evaluate overall forecasting performance under noisy conditions, we computed the average RMSE and MAE for all model variants within each architecture family—\autoformer, \informer, and \patchtst. Each model was assessed across five patch lengths and five forecast horizons for ten synthetic signals. The resulting heatmaps, shown in Figure~\ref{cross_family_comparison_noisy}, summarize the mean forecasting error of each model family across the full hyperparameter grid in the presence of input noise.

\paragraph{\autoformer~Family}
The \autoformer~family remains the most accurate under noisy conditions. RMSE values range from 0.046 to 0.076 across all patch and horizon combinations, while MAE values fall between 0.038 and 0.059. The model’s trend-seasonal decomposition continues to be effective, allowing it to filter out high-frequency noise and maintain predictive stability. Best results are observed for shorter forecast horizons (4 to 8 steps) and patch lengths between 12 and 20, where the model achieves an optimal balance between temporal context and noise resilience.

\paragraph{\patchtst~Family}
The \patchtst~family also performs well in noisy environments, with RMSE values generally below 0.07 and MAE under 0.058 in most settings. Its patch-based attention mechanism allows for effective modeling of local patterns, even when noise partially obscures signal features. Although there is a slight increase in error at longer horizons—particularly for \(H = 16\) and \(H = 20\)—the model maintains relatively stable performance across different patch lengths. This consistency makes \patchtst~a strong candidate for practical applications where robustness, efficiency, and training simplicity are key considerations.

\paragraph{\informer~Family}
The \informer~family experiences the largest decline in performance under noisy input. RMSE values reach up to 0.10 in several configurations, and MAE exceeds 0.06 consistently, peaking at 0.077 for long horizons and small patch lengths. These results indicate that the ProbSparse attention mechanism may struggle in noisy settings, where the approximation can omit important temporal dependencies. The heatmaps show high sensitivity to both patch length and horizon size, especially for configurations involving short patches and long-term predictions. This suggests that \informer~is less suited for forecasting tasks involving corrupted or noisy signals.

\paragraph{Patch and Horizon Sensitivity}
All three model families exhibit increased sensitivity to patch length and forecasting horizon when noise is present. The most reliable performance is achieved with patch lengths between 12 and 20 and forecast horizons ranging from 4 to 8. At shorter patch lengths, all models underperform due to limited contextual information. As the forecast horizon increases, error generally rises across all models; however, \autoformer~shows the smallest degradation, likely due to its decomposition mechanism. In contrast, \informer~shows steep declines in accuracy with increasing horizon length, particularly when paired with short patches.

In noisy forecasting scenarios, \autoformer~continues to offer the most accurate and stable performance. Its built-in trend-seasonal decomposition provides a strong inductive bias that helps isolate relevant patterns from noise. \patchtst~also performs well, showing reliable generalization across a variety of conditions and offering a computationally efficient alternative. The \informer~architecture, while fast and scalable, demonstrates greater vulnerability to noise, with larger prediction errors and less stable behavior across hyperparameter configurations. These findings emphasize the importance of architectural choices when working with noisy time series, highlighting the benefits of decomposition-based and locality-aware models.



\section{Outlook: Toward Koopman-Enhanced Transformer Architectures} \label{koopman}

The \textit{Koopman operator} provides a powerful, operator-theoretic framework to analyze nonlinear dynamical systems through the lens of linear dynamics in infinite-dimensional spaces. This operator acts on the space of observable functions, evolving them forward in time according to the system dynamics rather than directly propagating the states themselves \cite{Lusch2018,Yeung2019,NIPS2017_3a835d32}. This remarkable property enables the use of linear analysis techniques, even when the underlying system is nonlinear and potentially chaotic. Recent developments have revitalized interest in Koopman theory, particularly through data-driven methods such as Dynamic Mode Decomposition (DMD), which offer practical means to approximate the operator from measurements and facilitate modal analysis, prediction, and control in complex systems~\cite{Drgona2022,Skomski2021,NEURIPS2021_c9dd73f5}.

While Koopman-based methods have traditionally been applied to deterministic systems governed by ordinary or partial differential equations—such as fluid dynamics, robotics, and chaotic systems—they are increasingly being explored for data-driven forecasting. Prior works such as \cite{lusch2018deep}, \cite{takeishi2017learning} demonstrated how deep networks can be used to approximate Koopman-invariant subspaces, enabling stable rollout predictions over time. However, most of these studies focus on dynamical reconstruction or control, rather than general-purpose time series forecasting.

Our approach bridges this gap by combining Koopman operator learning with Transformer-based sequence models, such as \patchtst, \autoformer, and \informer. We use a Transformer encoder to extract temporally-aware latent representations from input time series patches, and then apply a constrained Koopman operator in the latent space to evolve the system forward. The operator is spectrally regularized to ensure stability, and the decoder projects the evolved latent states back into the original observable space. While classical forecasting methods treat time series as stochastic sequences with limited structural assumptions, our hybrid framework explicitly enforces a structured latent dynamic, making it well-suited for forecasting systems that arise from physical or biological processes. To the best of our knowledge, this work is among the first to integrate Koopman theory with Transformer-based backbones for robust, interpretable multi-step forecasting.

\subsection{Future Directions for Transformer-based Forecasting}
This article has already systematically benchmarked, analyzed, and proposed different variants for Transformer-based forecasting models, namely \patchtst, \informer, and \autoformer, across a wide range of synthetic signals and forecast settings. These models have demonstrated distinct advantages depending on signal characteristics, forecast horizons, and noise levels. However, all architectures primarily rely on direct autoregressive or encoder-decoder mechanisms, which, while effective, do not explicitly exploit the potential benefits of modeling system dynamics in latent linear spaces. Variants such as \informer, \autoformer, and \patchtst~are designed to address specific challenges like long-range dependencies, non-stationarity, and local pattern extraction. However, when modeling dynamical systems, such as oscillators or physical processes, simple prediction accuracy is not enough. These systems obey inherent physical principles, often expressed via: Stability properties (bounded energy, no explosion in latent space), Latent linear dynamics, Consistency over long horizons. 

\subsection{Deep Koopformer: Stability-Constrained Transformer-Koopman Framework}
An important direction for advancing time series forecasting lies in extending Transformer-based models—such as \patchtst, \informer, and \autoformer—by integrating them with principled representations of dynamical systems. To this end, we propose \emph{Deep Koopformer}, a hybrid framework that couples Transformer-based encoders with a stability-constrained Koopman operator. This design enables accurate forecasting while ensuring stable and interpretable latent dynamics, particularly for nonlinear or oscillatory systems.

The Koopman operator, a linear transformation acting on a lifted latent space, provides a natural bridge between deep sequence modeling and dynamical systems theory. By embedding the input sequence into a latent representation via a Transformer, and propagating this representation linearly through a parameterized Koopman operator, the Deep Koopformer enforces structured, stable latent evolution. This approach retains the flexibility of Transformers while introducing a physically motivated inductive bias into the temporal modeling process.

Traditional Transformer architectures, when optimized solely for short-term prediction accuracy, exhibit several well-known limitations:
\begin{itemize}
    \item \textit{Unstable latent dynamics:} Without explicit constraints, latent states may grow unbounded or collapse, leading to numerical instability.
    \item \textit{Forecast divergence:} Iterative rollouts amplify errors, especially in complex dynamical systems with sensitive dependence on initial conditions.
    \item \textit{Opaque latent space:} The lack of structure in the latent representation hinders interpretability and limits reliability in real-world settings.
\end{itemize}

By incorporating both \textit{spectral stability constraints} on the Koopman operator and a \textit{Lyapunov-inspired regularization} to penalize energy growth in latent space, Deep Koopformer directly addresses these challenges. The resulting model exhibits improved robustness, generalization, and alignment with the qualitative behavior of the underlying dynamical processes.


In the following section we detail different parts of the Deep Koopformer.


\subsubsection{Input-to-Output Transformation in Deep Koopformer}

The input consists of a sequence of observations:

\[
x_t \in \mathbb{R}^{d},
\]

where $x_t$ denotes the observed time series at time step $t$. This input is embedded into a latent space through a Transformer-based encoder:

\begin{equation}
z_t = \mathrm{Encoder}(x_t),
\end{equation}

where $z_t \in \mathbb{R}^{d_{\text{latent}}}$ is the latent representation learned to capture complex temporal dependencies.

The temporal evolution in the latent space is modeled via a learned Koopman operator $\mathcal{K}$, assumed linear:

\begin{equation}
z_{t+1} = \mathcal{K}(z_t).
\end{equation}

The Koopman operator is parametrized as:

\begin{equation}
\mathcal{K} = U \cdot \mathrm{diag}(S) \cdot V^\top,
\end{equation}

where $U$ and $V$ are orthogonal matrices and $S$ is a diagonal vector whose values are constrained to ensure stability:

\begin{equation}
S_i = \sigma(S_i^{\text{raw}}) \cdot 0.99.
\end{equation}

The propagated latent state is subsequently decoded to predict the next observed state:

\begin{equation}
\hat{x}_{t+1} = \mathrm{Decoder}(z_{t+1}),
\end{equation}

where $\hat{x}_{t+1} \in \mathbb{R}^{d}$ is the predicted observation at the next time step. The complete transformation can thus be expressed as:

\begin{equation}
\hat{x}_{t+1} = \mathrm{Decoder}(\mathcal{K}(\mathrm{Encoder}(x_t))).
\end{equation}

\vspace{1em}
\subsubsection{Spectral Stability via Koopman Parametrization}

In unconstrained models, latent dynamics may become unstable, leading to diverging predictions and poor generalization in long-horizon forecasting. To address this, we enforce spectral stability within the Koopman operator design. By construction, the spectral radius of $\mathcal{K}$ is strictly controlled:

\begin{equation}
\rho(\mathcal{K}) = \max |\lambda_i| \leq 0.99.
\end{equation}

This ensures that the latent dynamics are asymptotically stable, as repeated application of $\mathcal{K}$ satisfies:

\begin{equation}
\lim_{n \to \infty} \mathcal{K}^n = 0.
\end{equation}

As a result, the latent state trajectory remains bounded and avoids divergence during recursive forecasting, supporting stable and reliable predictions.

\vspace{1em}
\subsubsection{Energy Stability via Lyapunov Loss}

While spectral radius control limits eigenvalue magnitude, temporary growth in latent energy during training can still occur. Prediction loss alone does not penalize such undesirable latent behavior.

To further regulate latent dynamics, we introduce a Lyapunov-inspired loss function that penalizes increases in latent norm:

\begin{equation}
L_{\mathrm{Lyapunov}} = \mathrm{ReLU}(\|z_{t+1}\|^2 - \|z_t\|^2).
\end{equation}

This loss encourages contractive latent dynamics by penalizing instances where $z_{t+1}$ has higher energy than $z_t$. The use of ReLU ensures only positive differences contribute to the loss.

This energy-based regularization improves stability, prevents latent state explosion during training, and enhances robustness in noisy environments.

\vspace{1em}
\subsubsection{Orthogonality and Numerical Stability}

In order to guarantee the well-conditioning of the Koopman operator and avoid degenerate latent projections, orthogonality of the projection matrices $U$ and $V$ is enforced through Householder orthogonalization (Householder QR orthogonalization):
\begin{equation}
U^\top U = I, \quad V^\top V = I.
\end{equation}

Orthogonal matrices ensure that latent transformations remain rotation-like without amplifying or distorting latent representations.

This prevents ill-conditioning during optimization, improves numerical stability, and preserves interpretability of latent dynamics.

\vspace{1em}
\subsection{Koopman-Enhanced Transformer Formulation}

In summary, the proposed Deep Koopformer formulation combines Transformer-based encoders with stability-constrained Koopman operators, offering:

\begin{itemize}
    \item Flexible temporal modeling via Transformer backbones (\patchtst, \autoformer, \informer).
    \item Linear latent propagation with stability via Koopman operator and spectral constraints.
    \item Energy regulation through Lyapunov loss to avoid latent explosion.
    \item Numerical robustness through orthogonalization of Koopman matrices.
\end{itemize}

This framework ensures accurate, interpretable, and robust long-horizon forecasting while respecting the inherent physical properties of dynamical systems.


\subsection{Application to Noisy Dynamical Systems}

An important application area for Koopman-enhanced Transformer architectures is the forecasting of noisy nonlinear dynamical systems. As demonstrated in this study with synthetic signals, real-world processes often exhibit complex temporal dynamics combined with stochastic perturbations.

\paragraph{Van der Pol oscillator}

The Van der Pol oscillator is a second-order nonlinear ordinary differential equation (ODE) used to model various physical phenomena. The standard form of the Van der Pol equation is given by:

\[
\frac{d^2x}{dt^2} - \mu(1 - x^2)\frac{dx}{dt} + x = 0
\]

where:
\begin{itemize}
    \item \( x(t) \) is the state of the system (position),
    \item \( \frac{dx}{dt} \) is the velocity (first derivative),
    \item \( \frac{d^2x}{dt^2} \) is the acceleration (second derivative),
    \item \( \mu \) is a nonlinearity parameter that controls the amplitude and damping of the system.
\end{itemize}

This system can be rewritten as a set of two first-order differential equations by defining the state vector \( \mathbf{z}(t) = \begin{bmatrix} x_1(t) \\ x_2(t) \end{bmatrix} \), where \( x(t) \) is the position and \( v(t) = \frac{dx}{dt} \) is the velocity:

\[
\frac{d}{dt} \begin{bmatrix} x_1(t) \\ x_2(t) \end{bmatrix} = \begin{bmatrix} x_2(t) \\ \mu(1 - x_1(t)^2)x_2(t) - x_1(t) \end{bmatrix}
\]

Thus, the system of first-order equations governing the dynamics of the Van der Pol oscillator is:

\[
\frac{d}{dt} \mathbf{z}(t) = \mathbf{f}(\mathbf{z}(t)),
\]
where \( \mathbf{f}(\mathbf{z}(t)) \) is the nonlinear function defined as:

\[
\mathbf{f}(\mathbf{z}(t)) = \begin{bmatrix} x_2(t) \\ \mu(1 - x_1(t)^2)x_2(t) - x_1(t) \end{bmatrix}.
\]


In order to simulate realistic conditions and account for measurement noise or model uncertainties, stochastic perturbations are added to the system dynamics. The discretized version of the noisy Van der Pol oscillator can be written as:

\[
\mathbf{z}_{t+1} = \mathbf{z}_t + \Delta t \cdot \mathbf{f}(\mathbf{z}_t) + \boldsymbol{\eta}_t,
\]

where:
\begin{itemize}
    \item \( \mathbf{z}_t = \begin{bmatrix} x_{1,t} \\ x_{2,t} \end{bmatrix} \) is the state vector at time step \( t \),
    \item \( \Delta t \) is the discrete time step size,
    \item \( \mathbf{f}(\mathbf{z}_t) \) represents the deterministic Van der Pol dynamics,
    \item \( \boldsymbol{\eta}_t \sim \mathcal{N}(\mathbf{0}, \sigma^2 \mathbf{I}) \) is additive Gaussian noise with standard deviation \( \sigma \).
\end{itemize}

In this work, the noise term \( \boldsymbol{\eta}_t \) is sampled independently at each time step with standard deviation \( \sigma = 0.02 \), which introduces stochastic variability into both position and velocity.


The parameters used for the simulation are:
\begin{itemize}
    \item \( \mu = 1.0 \) (nonlinearity parameter),
    \item Time step: \( dt = 0.01 \),
    \item Total simulation time: \( T = 20.0 \) seconds,
    \item Initial conditions: \( x_1(0) = 2.0 \), \( x_2(0) = 0.0 \),
    \item Noise standard deviation: \( \sigma = 0.02 \).
\end{itemize}

The system is discretized into \( N = \frac{T}{dt} = 2000 \) time steps. The Van der Pol oscillator's noisy state trajectory is computed iteratively using the above update equation.

The evaluation of the Deep Koopformer models focuses on forecasting the next \(H= 5\) time steps of the system's dynamics. The models are trained using input sequences of length \(16\) (referred to as the \textit{patch length}), with the objective of predicting the future values of two state variables, \(x_1\) and \(x_2\), over the forecast horizon. The target outputs consist of \(10\) variables, representing the forecasted values of the state variables for each of the \(5\) time steps. These models are evaluated by comparing the predicted outputs with the ground truth, assessing both short-term accuracy and long-term stability.

The training process is carried out with a consistent set of hyperparameters across all variants of the Deep Koopformer architecture (\texttt{PatchTST}, \texttt{Autoformer}, and \texttt{Informer}). The latent dimension of the model is set to \(d_\text{model} = 16\), which determines the size of the internal representation of the state dynamics. The multi-head attention mechanism is configured with \(2\) attention heads, enabling the model to capture multiple dependencies in the temporal sequence. The feed-forward dimension is \(64\), which defines the size of the intermediate layer within the Transformer encoder, while the number of encoder layers is set to \(2\), specifying the depth of the encoder stack. 

The Adam optimizer with a learning rate of \(0.001\) is used for gradient-based optimization. The loss function employed is the mean squared error (MSE), augmented with a stability penalty that helps ensure the preservation of latent dynamics over time. The model is trained for a total of \(1000\) epochs, which provides sufficient training time for the model to learn the complex temporal dependencies of the system and accurately forecast future states. The training process includes regular evaluation of the model to ensure convergence and stability in the learned dynamics.

\begin{figure}
    \centering
    \includegraphics[width=0.99\linewidth]{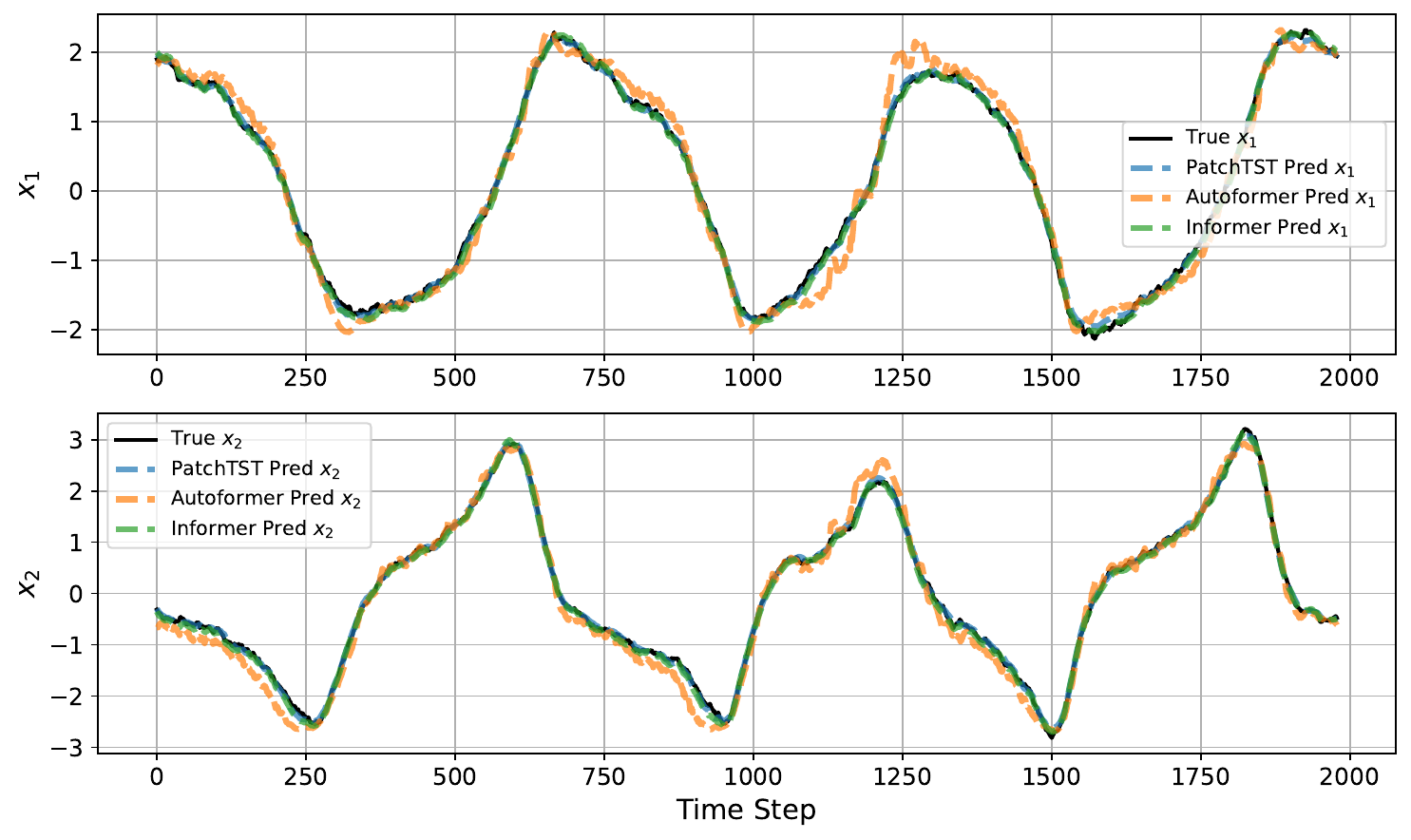}
    \caption{Deep Koopformer based Transformers Variants Comparison on the Van der Pol System.}
    \label{fig:van_der_pol}
\end{figure}

\begin{figure}
    \centering
    \includegraphics[width=0.75\linewidth]{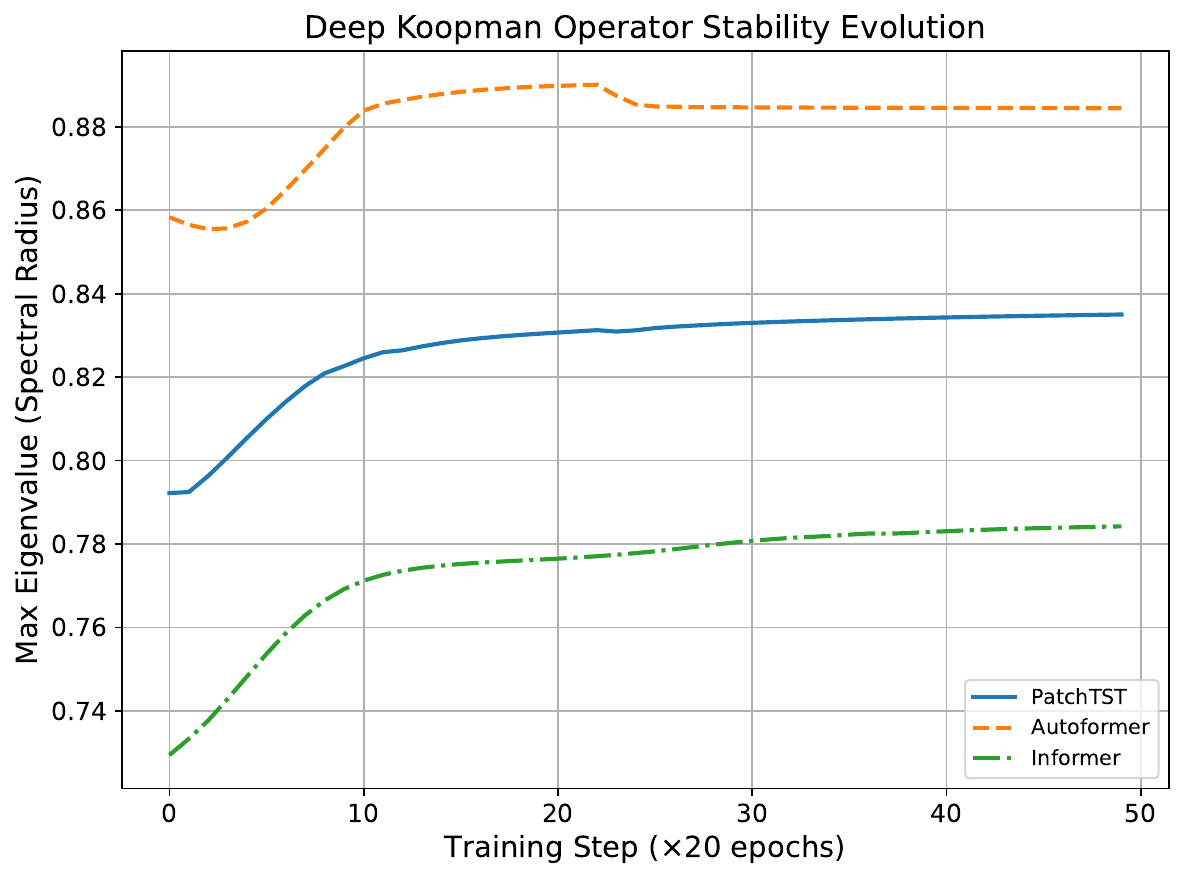}
    \caption{Comparison of Deep Koopman Operator Stability Evolution in Variants Transformers on Van der Pol System during the training loop.}
    \label{fig:van_der_pol_eigenvalue}
\end{figure}

As shown in Fig.~\ref{fig:van_der_pol} and Fig.~\ref{fig:van_der_pol_eigenvalue}, the results from the training and evaluation indicate that all variants of the Deep Koopformer architecture perform similarly well in capturing the nonlinear oscillatory behavior of the Van der Pol system. They achieve accurate forecasting of both state variables, \(x_1\) and \(x_2\), while maintaining stable latent dynamics. This stability is further ensured by the Koopman operator, which constrains the evolution of the latent states, and is validated through the spectral radius of the operator during training (see Fig.~\ref{fig:van_der_pol_eigenvalue}). 



\paragraph{Lorenz System}

The Lorenz system is a classic example of a chaotic dynamical system originally developed to model atmospheric convection. It consists of three coupled, nonlinear ordinary differential equations (ODEs) defined as:

$$
\begin{aligned}
\frac{dx_1}{dt} &= \sigma (x_2 - x_1), \\
\frac{dx_2}{dt} &= x_1 (\rho - x_3) - x_2, \\
\frac{dx_3}{dt} &= x_1 x_2 - \beta x_3,
\end{aligned}
$$

where:
\begin{itemize}
\item $x_1(t)$, $x_2(t)$, and $x_3(t)$ represent the state variables at time $t$,
\item $\sigma$ is the Prandtl number and controls the rate of convection,
\item $\rho$ is the Rayleigh number and governs the temperature difference,
\item $\beta$ is a geometric factor related to the aspect ratio of the convection cells.
\end{itemize}


To simulate more realistic conditions, stochastic noise is added to the deterministic Lorenz dynamics. The discretized and noisy Lorenz system is defined as:

$$
\mathbf{x}_{t+1} = \mathbf{x}_t + \Delta t \cdot \mathbf{g}(\mathbf{x}_t) + \boldsymbol{\eta}_t,
$$

where:
\begin{itemize}
\item $\mathbf{x}_t = \begin{bmatrix} x_{1,t} \\ x_{2,t} \\ x_{3,t} \end{bmatrix}$ is the state vector at time step $t$,
\item $\Delta t$ is the time step size,
\item $\mathbf{g}(\mathbf{x}_t)$ represents the deterministic Lorenz dynamics,
\item $\boldsymbol{\eta}_t \sim \mathcal{N}(\mathbf{0}, \sigma^2 \mathbf{I})$ is an additive Gaussian noise term.
\end{itemize}

The deterministic dynamics $\mathbf{g}(\mathbf{x}_t)$ are explicitly given by:

$$
\mathbf{g}(\mathbf{x}_t) = \begin{bmatrix}
\sigma (x_{2,t} - x_{1,t}) \\
x_{1,t} (\rho - x_{3,t}) - x_{2,t} \\
x_{1,t} x_{2,t} - \beta x_{3,t}
\end{bmatrix}.
$$


The parameters used for the Lorenz simulation are:
\begin{itemize}
\item $\sigma = 10.0$ (Prandtl number),
\item $\rho = 28.0$ (Rayleigh number),
\item $\beta = \frac{8}{3}$ (geometric factor),
\item Time step: $dt = 0.01$,
\item Total simulation time: $T = 20.0$ seconds,
\item Initial conditions: $x_1(0) = 1.0$, $x_2(0) = 1.0$, $x_3(0) = 1.0$,
\item Noise standard deviation: $\sigma = 0.5$.
\end{itemize}

The system is discretized into $N = \frac{T}{dt} = 2000$ time steps. The noisy trajectory is simulated by applying the above update equations iteratively with Gaussian noise added at each step to reflect random perturbations in the system.




The training of the Deep Koopformer \patchtst~Minimal backbone was conducted using patch length \(200\), with horizon of \(H= 5\) future time steps. The target outputs consist of \(15\) variables, corresponding to the predicted values of the system's three state variables (\(x_1\), \(x_2\), \(x_3\)) over the forecast horizon. The models were trained with a consistent set of hyperparameters: the latent dimension \(d_\text{model} = 16\), \(2\) attention heads, a feed-forward dimension of \(64\), and \(2\) encoder layers. The Adam optimizer with a learning rate of \(0.001\) was used for optimization, while the loss function was a combination of MSE and a stability penalty term based on the Lyapunov criterion. The model was trained for a total of \(3000\) epochs to ensure thorough learning of the temporal dynamics.

\begin{figure}
    \centering
    \includegraphics[width=0.99\linewidth]{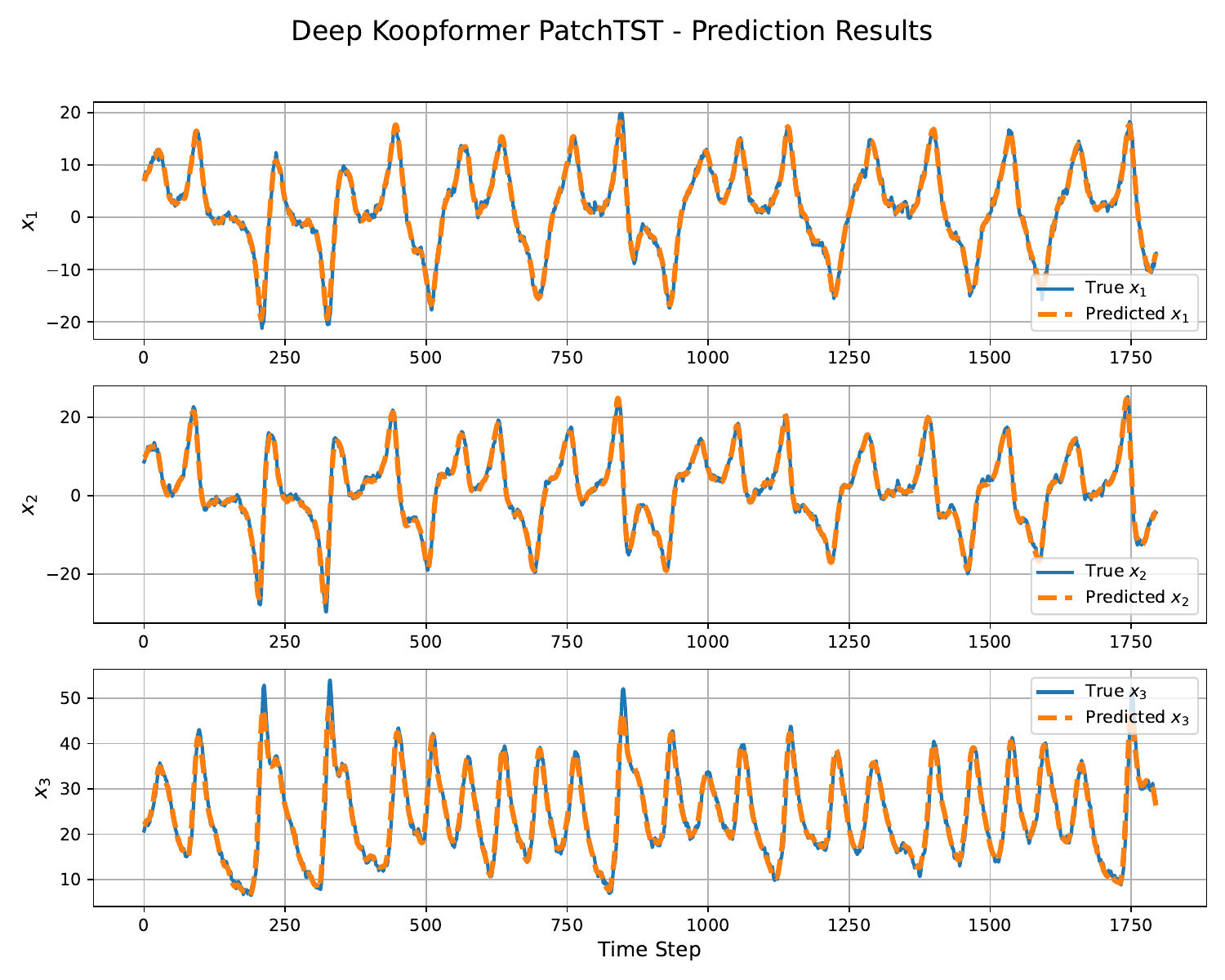}
    \caption{Deep Koopformer \patchtst prediction on Lorenz Dynamical System}
    \label{fig:lorenz_system}
\end{figure}

\begin{figure}
    \centering
    \includegraphics[width=0.75\linewidth]{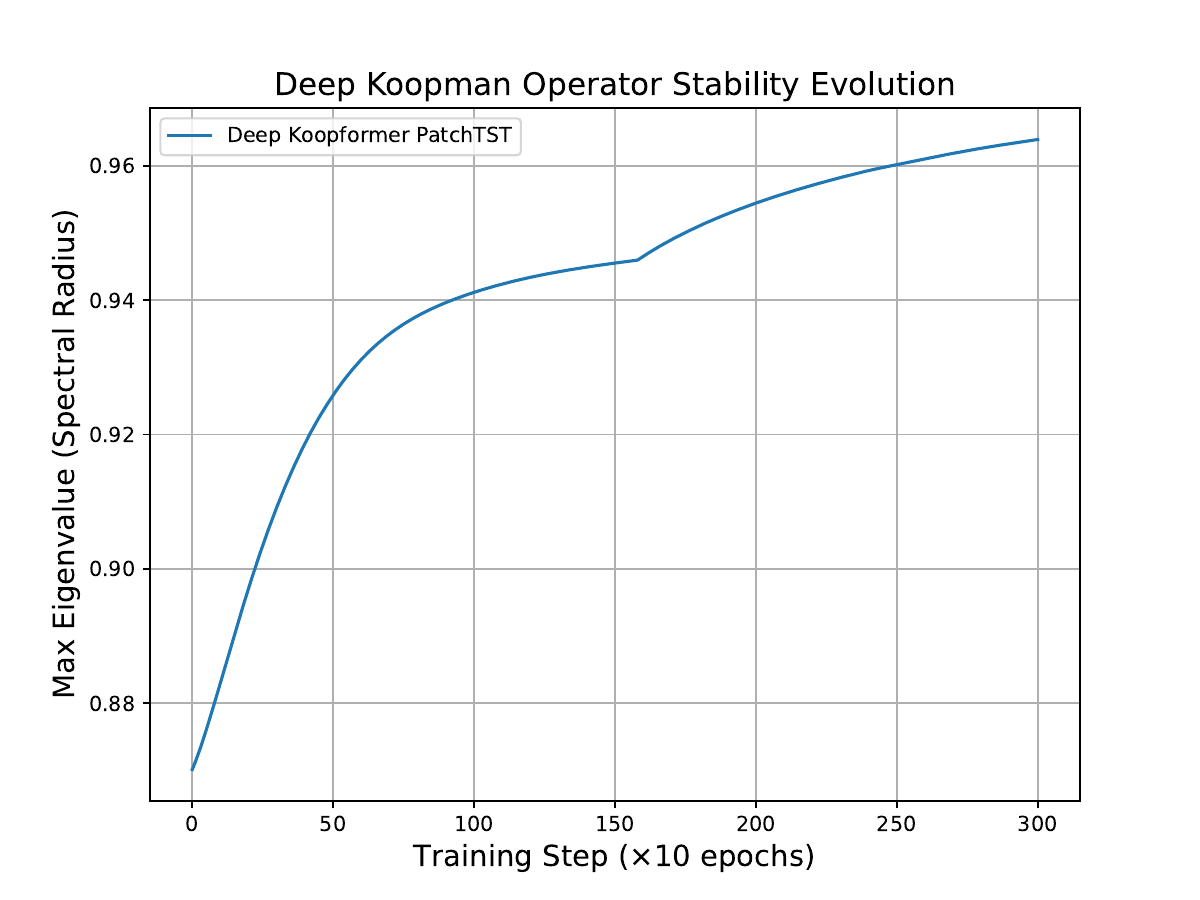}
    \caption{Deep Koopman Operator Stability Evolution with \patchtst~backbone during the training loop}
    \label{fig:lorenz_system_eigenvalue}
\end{figure}

The results shown in Fig. \ref{fig:lorenz_system} and Fig. \ref{fig:lorenz_system_eigenvalue} highlight the performance of the Deep Koopformer with \patchtst~Minimal backbone in predicting the dynamics of the Lorenz system. As shown in Fig. \ref{fig:lorenz_system}, the model effectively captures the oscillatory behavior of all three state variables (\(x_1\), \(x_2\), and \(x_3\)), with predicted values closely following the true values across the entire time series. The blue lines represent the true values, while the orange dashed lines represent the predicted values. The prediction results for \(x_1\), \(x_2\), and \(x_3\) demonstrate the model's capability to replicate the nonlinear dynamics of the system.

In Fig. \ref{fig:lorenz_system_eigenvalue}, the evolution of the Koopman operator's spectral radius is plotted over the course of training. The plot indicates that the model achieves stability in the latent space as the training progresses, with the maximum eigenvalue gradually increasing and stabilizing after the initial training steps. This stability is crucial for ensuring the model's robustness in long-term forecasting, confirming that the Koopman operator enforces stable latent dynamics throughout the training process.

\section{Conclusion} \label{conclusion_sec}
This study presents a unified empirical investigation of Transformer-based model families for univariate time series forecasting under both clean and noisy signal conditions. We benchmark three prominent architectures—\autoformer, \informer, and \patchtst~—each implemented in three variants: \textit{Minimal}, \textit{Standard}, and \textit{Full}, representing increasing levels of architectural complexity.

Across 10 synthetic signals and 5$\times$5 configurations of patch lengths and forecast horizons, we conduct 750 experiments per noise regime, totaling 1500 evaluations for each model family. This dual-setting analysis (clean + noisy) enables robust insights into the forecasting capabilities, stability, and noise resilience of each architecture.

\patchtst~Standard emerges as the most consistent and robust variant across both clean and noisy environments. It delivers low RMSE and MAE even under signal perturbations, demonstrating resilience in long-horizon, non-stationary, and noisy conditions while remaining computationally efficient. Patch-based attention with learned temporal embeddings proves highly effective.

\autoformer~Minimal and \autoformer~Standard maintain competitive accuracy, particularly in clean, short-to-medium horizon tasks. Their performance is bolstered by trend-seasonal decomposition, which imparts stability and interpretability. Notably, \autoformer~continues to outperform in noisy settings—confirming its structural advantage in capturing underlying signal trends even amidst distortion.

\informer~Standard and \informer~Full exhibit clear performance degradation under noise. While sparse attention offers scalability, it leads to unstable forecasts, especially at longer horizons and with larger patch sizes. \informer~struggles most with noisy inputs, indicating its limited generalization in structured low-noise or high-variance time series.

For all three model families, performance is sensitive to both forecast horizon and patch length. Optimal results are generally found with patch lengths in the range of 12–16 and forecast horizons of 2–6. Increasing patch size beyond this range does not reliably improve accuracy and may induce overfitting, particularly in clean settings.

To advance beyond these autoregressive and encoder-decoder designs, we further introduced the Koopman-enhanced Transformer architecture (\textit{Koopformer}). By integrating Koopman operator theory to model linear latent state evolution, Koopformer provides a hybrid framework that unifies deep sequence modeling with dynamical systems theory. Empirical results on nonlinear and chaotic systems, including the noisy Van der Pol oscillator and Lorenz attractor, demonstrate that Koopformer variants can effectively capture nonlinear oscillations and chaotic trajectories with improved stability and interpretability.

In conclusion, encoder-only models such as \patchtst~Standard offer the best trade-off between forecasting accuracy, robustness, and scalability across noise regimes. Trend decomposition of \autoformer~remains beneficial under distortion, while \informer~is less suited to structured and noisy time series. Koopman-enhanced Transformer expands the design space, offering a promising hybrid approach for interpretable and stable forecasting of nonlinear dynamical systems. Future work will explore extending Koopman enahnced Transformer to multivariate and real-world datasets, and integrating advanced operator learning and interpretability mechanisms to further enhance robustness and scientific insight.

\appendix


\begin{algorithm}[!t]
\caption{\patchtst~Full Forecasting Procedure}
\label{alg:patchtstfull}
\begin{algorithmic}[1]
\State \textbf{Input:} Patch sequence $\mathbf{x} \in \mathbb{R}^{B \times P \times 1}$; forecast horizon $H$; model dimension $d_{\text{model}}$
\State \textbf{Output:} Forecast $\hat{\mathbf{y}} \in \mathbb{R}^{B \times H}$
\State \textbf{Step 1: Encoder Input Projection and Positional Encoding}
\[
\mathbf{Z}^{\text{enc}} = \mathbf{x} \cdot \mathbf{W}_e^{(e)} + \text{PE}^{(e)} \in \mathbb{R}^{B \times P \times d_{\text{model}}}
\]
\State \textbf{Step 2: Transformer Encoder Layers}
\For{each encoder layer $\ell = 1, \dots, L$}
    \State Compute attention heads:
    \[
    Q = \mathbf{Z}^{\text{enc}} \cdot \mathbf{W}^{Q}_\ell,\quad
    K = \mathbf{Z}^{\text{enc}} \cdot \mathbf{W}^{K}_\ell,\quad
    V = \mathbf{Z}^{\text{enc}} \cdot \mathbf{W}^{V}_\ell
    \]
    \[
    \text{head}_i = \text{Softmax}\left( \frac{Q_i K_i^\top}{\sqrt{d_{\text{model}}/h}} \right) V_i
    \]
    \[
    \text{MHAttn} = \text{Concat}(\text{head}_1, \dots, \text{head}_h) \cdot \mathbf{W}^{O}_\ell
    \]
    \State Apply residual connection and layer normalization:
    \[
    \mathbf{Z}^{\text{enc}} \gets \text{LayerNorm}(\mathbf{Z}^{\text{enc}} + \text{MHAttn})
    \]
    \State Apply feedforward and second residual block:
    \[
    \mathbf{Z}^{\text{enc}} \gets \text{LayerNorm}(\mathbf{Z}^{\text{enc}} + \text{FFN}(\mathbf{Z}^{\text{enc}}))
    \]
\EndFor
\State Encoder output: $\mathbf{H}^{\text{enc}} = \mathbf{Z}^{\text{enc}}$

\State \textbf{Step 3: Decoder Input Construction and Projection}
\State Repeat the final time step $H$ times:
\[
\mathbf{x}^{\text{rep}} = \text{Repeat}(\mathbf{x}_{:, -1, :}, H) \in \mathbb{R}^{B \times H \times 1}
\]
\State Project and encode with positional embedding:
\[
\mathbf{Z}^{\text{dec}} = \mathbf{x}^{\text{rep}} \cdot \mathbf{W}_e^{(d)} + \text{PE}^{(d)} \in \mathbb{R}^{B \times H \times d_{\text{model}}}
\]

\State \textbf{Step 4: Transformer Decoder Layer}
\State Compute self-attention on decoder input:
\[
Q = \mathbf{Z}^{\text{dec}} \cdot \mathbf{W}^{Q_d},\quad
K = \mathbf{Z}^{\text{dec}} \cdot \mathbf{W}^{K_d},\quad
V = \mathbf{Z}^{\text{dec}} \cdot \mathbf{W}^{V_d}
\]
\[
\text{SelfAttn} = \text{Softmax}\left( \frac{Q K^\top}{\sqrt{d_{\text{model}}}} \right) V
\]

\State Compute encoder-decoder cross-attention:
\[
Q = \mathbf{Z}^{\text{dec}} \cdot \mathbf{W}^{Q_c},\quad
K = \mathbf{H}^{\text{enc}} \cdot \mathbf{W}^{K_c},\quad
V = \mathbf{H}^{\text{enc}} \cdot \mathbf{W}^{V_c}
\]
\[
\text{CrossAttn} = \text{Softmax}\left( \frac{Q K^\top}{\sqrt{d_{\text{model}}}} \right) V
\]

\State Apply decoder residual blocks:
\[
\mathbf{Z}^{\text{dec}} \gets \text{LayerNorm}(\mathbf{Z}^{\text{dec}} + \text{SelfAttn})
\]
\[
\mathbf{Z}^{\text{dec}} \gets \text{LayerNorm}(\mathbf{Z}^{\text{dec}} + \text{CrossAttn})
\]
\[
\mathbf{Z}^{\text{dec}} \gets \text{LayerNorm}(\mathbf{Z}^{\text{dec}} + \text{FFN}(\mathbf{Z}^{\text{dec}}))
\]

\State \textbf{Step 5: Output Projection}
\[
\hat{\mathbf{y}} = \mathbf{Z}^{\text{dec}} \cdot \mathbf{W}_o \in \mathbb{R}^{B \times H \times 1}
\]
\State \Return $\hat{\mathbf{y}}$ reshaped to $\mathbb{R}^{B \times H}$
\end{algorithmic}
\end{algorithm}


\begin{algorithm}[!h]
\caption{Full \informer~ Forecasting Procedure with ProbSparse Attention}
\label{alg:fullinformer}
\begin{algorithmic}[1]
\State \textbf{Input:} Input sequence $\mathbf{x} \in \mathbb{R}^{B \times P \times 1}$, forecast horizon $H$, model dimension $d_{\text{model}}$
\State \textbf{Output:} Forecast sequence $\hat{\mathbf{y}} \in \mathbb{R}^{B \times H}$
\State \textbf{Step 1: Encoder Projection and Positional Encoding}
\[
\mathbf{Z}^{\text{enc}} = \mathbf{x} \cdot \mathbf{W}_e^{(e)} + \text{PE}^{(e)} \in \mathbb{R}^{B \times P \times d_{\text{model}}}
\]

\State \textbf{Step 2: ProbSparse Transformer Encoder}
\For{each encoder layer}
    \State Compute attention triplets:
    \[
    Q = \mathbf{Z}^{\text{enc}} \cdot \mathbf{W}^Q, \quad 
    K = \mathbf{Z}^{\text{enc}} \cdot \mathbf{W}^K, \quad 
    V = \mathbf{Z}^{\text{enc}} \cdot \mathbf{W}^V
    \]
    \State Compute sparsity score for each $q_i$:
    \[
    M(q_i) = \max_j \left( \frac{q_i k_j^\top}{\sqrt{d_{\text{model}}}} \right) - \frac{1}{P} \sum_{j=1}^{P} \frac{q_i k_j^\top}{\sqrt{d_{\text{model}}}}
    \]
    \State Select top-$u$ queries ($u = c \log P$) based on $M(q_i)$
    \State For top-$u$ queries, compute full attention:
    \[
    \text{Attn}(q_i, K, V) = \text{Softmax} \left( \frac{q_i K^\top}{\sqrt{d_{\text{model}}}} \right) V
    \]
    \State For other queries, use top-$u$ keys (sparse approximation)
    \State Apply feedforward layers and residual connections
\EndFor
\State Let encoder output: $\mathbf{H}^{\text{enc}} \in \mathbb{R}^{B \times P \times d_{\text{model}}}$

\State \textbf{Step 3: Decoder Initialization}
\State Repeat last input value $H$ times:
\[
\mathbf{x}^{\text{rep}} = \text{Repeat}(\mathbf{x}_{:, -1, :}, H) \in \mathbb{R}^{B \times H \times 1}
\]
\State Project and add positional encoding:
\[
\mathbf{Z}^{\text{dec}} = \mathbf{x}^{\text{rep}} \cdot \mathbf{W}_e^{(d)} + \text{PE}^{(d)} \in \mathbb{R}^{B \times H \times d_{\text{model}}}
\]

\State \textbf{Step 4: ProbSparse Transformer Decoder}
\For{each decoder layer}
    \State Compute self-attention using ProbSparse:
    \[
    Q = \mathbf{Z}^{\text{dec}} \cdot \mathbf{W}^{Q_d}, \quad 
    K = \mathbf{Z}^{\text{dec}} \cdot \mathbf{W}^{K_d}, \quad 
    V = \mathbf{Z}^{\text{dec}} \cdot \mathbf{W}^{V_d}
    \]
    \[
    \text{SelfAttn} = \text{SparseAttention}(Q, K, V)
    \]
    \State Compute cross-attention with encoder output:
    \[
    Q = \mathbf{Z}^{\text{dec}} \cdot \mathbf{W}^{Q_c}, \quad 
    K = \mathbf{H}^{\text{enc}} \cdot \mathbf{W}^{K_c}, \quad 
    V = \mathbf{H}^{\text{enc}} \cdot \mathbf{W}^{V_c}
    \]
    \[
    \text{CrossAttn} = \text{Softmax}\left( \frac{Q K^\top}{\sqrt{d_{\text{model}}}} \right) V
    \]
    \State Apply residual and feedforward layers
\EndFor
\State Let decoder output: $\mathbf{H}^{\text{dec}} \in \mathbb{R}^{B \times H \times d_{\text{model}}}$

\State \textbf{Step 5: Output Projection}
\[
\hat{\mathbf{y}} = \mathbf{H}^{\text{dec}} \cdot \mathbf{W}_o \in \mathbb{R}^{B \times H \times 1}
\]
\State \Return $\hat{\mathbf{y}}$ (squeezed to shape $\mathbb{R}^{B \times H}$)
\end{algorithmic}
\end{algorithm}


\begin{algorithm}[!h]
\caption{Full \autoformer~Forecasting Procedure}
\label{alg:fullautoformer}
\begin{algorithmic}[1]
\State \textbf{Input:} Sequence $\mathbf{x} \in \mathbb{R}^{B \times P \times 1}$, forecast horizon $H$, model dimension $d_{\text{model}}$
\State \textbf{Output:} Forecast $\hat{\mathbf{y}} \in \mathbb{R}^{B \times H}$

\State \textbf{Step 1: Series Decomposition}
\State Apply moving average of kernel size $k$:
\[
\mathbf{x}^{\text{trend}} = \text{MA}_k(\mathbf{x}), \quad \mathbf{x}^{\text{seasonal}} = \mathbf{x} - \mathbf{x}^{\text{trend}}
\]

\State \textbf{Step 2: Encoder Path}
\State Project seasonal input:
\[
\mathbf{Z}^{\text{enc}} = \mathbf{x}^{\text{seasonal}} \cdot \mathbf{W}_e^{(e)} + \text{PE}^{(e)} \in \mathbb{R}^{B \times P \times d_{\text{model}}}
\]
\State Apply $L$ encoder layers (self-attention + FFN + residual + norm). For each layer:
\[
Q = \mathbf{Z}^{\text{enc}} \cdot \mathbf{W}^Q, \quad 
K = \mathbf{Z}^{\text{enc}} \cdot \mathbf{W}^K, \quad 
V = \mathbf{Z}^{\text{enc}} \cdot \mathbf{W}^V
\]
\[
\text{SelfAttn}(Q, K, V) = \text{Softmax}\left( \frac{Q K^\top}{\sqrt{d_{\text{model}}}} \right) V
\]
\State Let encoder output: $\mathbf{H}^{\text{enc}} \in \mathbb{R}^{B \times P \times d_{\text{model}}}$

\State \textbf{Step 3: Decoder Initialization}
\State Zero initialize seasonal decoder input:
\[
\mathbf{z}^{\text{dec}}_0 = \mathbf{0} \in \mathbb{R}^{B \times H \times 1}
\]
\State Project decoder input:
\[
\mathbf{Z}^{\text{dec}} = \mathbf{z}^{\text{dec}}_0 \cdot \mathbf{W}_e^{(d)} + \text{PE}^{(d)} \in \mathbb{R}^{B \times H \times d_{\text{model}}}
\]

\State \textbf{Step 4: Transformer Decoder}
\For{each decoder layer}
    \State \textit{(a) Self-Attention:}
    \[
    Q = K = V = \mathbf{Z}^{\text{dec}} \cdot \mathbf{W}^{Q_d}, \mathbf{W}^{K_d}, \mathbf{W}^{V_d}
    \]
    \[
    \text{SelfAttn} = \text{Softmax} \left( \frac{Q K^\top}{\sqrt{d_{\text{model}}}} \right) V
    \]
    
    \State \textit{(b) Cross-Attention:}
    \[
    Q = \mathbf{Z}^{\text{dec}} \cdot \mathbf{W}^{Q_c}, \quad 
    K = \mathbf{H}^{\text{enc}} \cdot \mathbf{W}^{K_c}, \quad 
    V = \mathbf{H}^{\text{enc}} \cdot \mathbf{W}^{V_c}
    \]
    \[
    \text{CrossAttn} = \text{Softmax} \left( \frac{Q K^\top}{\sqrt{d_{\text{model}}}} \right) V
    \]
    
    \State Apply FFN, residuals, and normalization
\EndFor
\State Let decoder output: $\mathbf{H}^{\text{dec}} \in \mathbb{R}^{B \times H \times d_{\text{model}}}$

\State \textbf{Step 5: Forecast Projection}
\State Forecast seasonal output:
\[
\hat{\mathbf{y}}^{\text{seasonal}} = \mathbf{H}^{\text{dec}} \cdot \mathbf{W}_o \in \mathbb{R}^{B \times H}
\]
\State Forecast trend output:
\[
\hat{\mathbf{y}}^{\text{trend}} = \mathbf{x}^{\text{trend}} \cdot \mathbf{W}_t \in \mathbb{R}^{B \times H}
\]

\State \textbf{Step 6: Final Output}
\[
\hat{\mathbf{y}} = \hat{\mathbf{y}}^{\text{seasonal}} + \hat{\mathbf{y}}^{\text{trend}} \in \mathbb{R}^{B \times H}
\]
\State \Return $\hat{\mathbf{y}}$
\end{algorithmic}
\end{algorithm}


\bibliographystyle{ieeetr}

\bibliography{my_ref,my_ref_2}

\begin{thebibliography}{10}

\bibitem{forootani2024climate}
A.~Forootani, D.~E. Aliabadi, and D.~Thraen, ``Climate aware deep neural
  networks (cadnn) for wind power simulation,'' {\em arXiv preprint
  arXiv:2412.12160}, 2024.

\bibitem{alvarez2010energy}
F.~M. Alvarez, A.~Troncoso, J.~C. Riquelme, and J.~S.~A. Ruiz, ``Energy time
  series forecasting based on pattern sequence similarity,'' {\em IEEE
  Transactions on Knowledge and Data Engineering}, vol.~23, no.~8,
  pp.~1230--1243, 2010.

\bibitem{dingli2017financial}
A.~Dingli and K.~S. Fournier, ``Financial time series forecasting-a deep
  learning approach,'' {\em International Journal of Machine Learning and
  Computing}, vol.~7, no.~5, pp.~118--122, 2017.

\bibitem{wang2024unveiling}
S.~Wang, Y.~Lin, Y.~Jia, J.~Sun, and Z.~Yang, ``Unveiling the multi-dimensional
  spatio-temporal fusion transformer (mdstft): A revolutionary deep learning
  framework for enhanced multi-variate time series forecasting,'' {\em IEEE
  Access}, 2024.

\bibitem{yan2024multi}
K.~Yan, C.~Long, H.~Wu, and Z.~Wen, ``Multi-resolution expansion of analysis in
  time-frequency domain for time series forecasting,'' {\em IEEE Transactions
  on Knowledge and Data Engineering}, 2024.

\bibitem{7438940}
A.~Tascikaraoglu, B.~M. Sanandaji, G.~Chicco, V.~Cocina, F.~Spertino,
  O.~Erdinc, N.~G. Paterakis, and J.~P. Catalão, ``Compressive spatio-temporal
  forecasting of meteorological quantities and photovoltaic power,'' {\em IEEE
  Transactions on Sustainable Energy}, vol.~7, no.~3, pp.~1295--1305, 2016.

\bibitem{benidis2022deep}
K.~Benidis, S.~S. Rangapuram, V.~Flunkert, Y.~Wang, D.~Maddix, C.~Turkmen,
  J.~Gasthaus, M.~Bohlke-Schneider, D.~Salinas, L.~Stella, {\em et~al.}, ``Deep
  learning for time series forecasting: Tutorial and literature survey,'' {\em
  ACM Computing Surveys}, vol.~55, no.~6, pp.~1--36, 2022.

\bibitem{hewage2020temporal}
P.~Hewage, A.~Behera, M.~Trovati, E.~Pereira, M.~Ghahremani, F.~Palmieri, and
  Y.~Liu, ``Temporal convolutional neural (tcn) network for an effective
  weather forecasting using time-series data from the local weather station,''
  {\em Soft Computing}, vol.~24, pp.~16453--16482, 2020.

\bibitem{survey1}
L.~Bryan and Z.~Stefan, ``Time-series forecasting with deep learning: a
  survey,'' {\em Phil. Trans. R. Soc. A}, 2021.

\bibitem{survey2}
J.~F. Torres, D.~Hadjout, A.~Sebaa, F.~Mart{\'\i}nez-{\'A}lvarez, and
  A.~Troncoso, ``Deep learning for time series forecasting: a survey,'' {\em
  Big Data}, vol.~9, no.~1, pp.~3--21, 2021.

\bibitem{survey3}
P.~Lara-Ben{\'\i}tez, M.~Carranza-Garc{\'\i}a, and J.~C. Riquelme, ``An
  experimental review on deep learning architectures for time series
  forecasting,'' {\em International Journal of Neural Systems}, vol.~31,
  no.~03, p.~2130001, 2021.

\bibitem{nlpsurvey}
K.~S. Kalyan, A.~Rajasekharan, and S.~Sangeetha, ``Ammus: A survey of
  transformer-based pretrained models in natural language processing,'' {\em
  arXiv preprint arXiv:2108.05542}, 2021.

\bibitem{transformer}
A.~Vaswani, N.~Shazeer, N.~Parmar, J.~Uszkoreit, L.~Jones, A.~N. Gomez,
  L.~Kaiser, and I.~Polosukhin, ``Attention is all you need,'' in {\em Advances
  in Neural Information Processing Systems}, vol.~30, 2017.

\bibitem{cvsurvey}
S.~Khan, M.~Naseer, M.~Hayat, S.~W. Zamir, F.~S. Khan, and M.~Shah,
  ``Transformers in vision: A survey,'' {\em ACM Computing Surveys (CSUR)},
  2021.

\bibitem{speechsurvey}
S.~Karita, N.~Chen, T.~Hayashi, T.~Hori, H.~Inaguma, Z.~Jiang, M.~Someki,
  N.~E.~Y. Soplin, R.~Yamamoto, X.~Wang, {\em et~al.}, ``A comparative study on
  transformer vs rnn in speech applications,'' in {\em IEEE Automatic Speech
  Recognition and Understanding Workshop (ASRU)}, pp.~449--456, IEEE, 2019.

\bibitem{tssurvey}
Q.~Wen, T.~Zhou, C.~Zhang, W.~Chen, Z.~Ma, J.~Yan, and L.~Sun, ``Transformers
  in time series: A survey,'' {\em arXiv preprint arXiv:2202.07125}, 2022.

\bibitem{fournier2023practical}
Q.~Fournier, G.~M. Caron, and D.~Aloise, ``A practical survey on faster and
  lighter transformers,'' {\em ACM Computing Surveys}, vol.~55, no.~14s,
  pp.~1--40, 2023.

\bibitem{zhu2021long}
C.~Zhu, W.~Ping, C.~Xiao, M.~Shoeybi, T.~Goldstein, A.~Anandkumar, and
  B.~Catanzaro, ``Long-short transformer: Efficient transformers for language
  and vision,'' {\em Advances in neural information processing systems},
  vol.~34, pp.~17723--17736, 2021.

\bibitem{informer}
H.~Zhou, S.~Zhang, J.~Peng, S.~Zhang, J.~Li, H.~Xiong, and W.~Zhang,
  ``Informer: Beyond efficient transformer for long sequence time-series
  forecasting,'' in {\em The Thirty-Fifth {AAAI} Conference on Artificial
  Intelligence}, vol.~35, pp.~11106--11115, 2021.

\bibitem{autoformer}
H.~Wu, J.~Xu, J.~Wang, and M.~Long, ``Autoformer: Decomposition transformers
  with {Auto-Correlation} for long-term series forecasting,'' in {\em Advances
  in Neural Information Processing Systems}, 2021.

\bibitem{fedformer}
T.~Zhou, Z.~Ma, Q.~Wen, X.~Wang, L.~Sun, and R.~Jin, ``{FEDformer}: Frequency
  enhanced decomposed transformer for long-term series forecasting,'' in {\em
  Proc. 39th International Conference on Machine Learning}, 2022.

\bibitem{pyraformer}
S.~Liu, H.~Yu, C.~Liao, J.~Li, W.~Lin, A.~X. Liu, and S.~Dustdar, ``Pyraformer:
  Low-complexity pyramidal attention for long-range time series modeling and
  forecasting,'' in {\em International Conference on Learning Representations},
  2022.

\bibitem{dlinear}
A.~Zeng, M.~Chen, L.~Zhang, and Q.~Xu, ``Are transformers effective for time
  series forecasting?,'' {\em arXiv preprint arXiv:2205.13504}, 2022.

\bibitem{nie2022time}
Y.~Nie, N.~H. Nguyen, P.~Sinthong, and J.~Kalagnanam, ``A time series is worth
  64 words: Long-term forecasting with transformers,'' {\em arXiv preprint
  arXiv:2211.14730}, 2022.

\bibitem{vit}
A.~Dosovitskiy, L.~Beyer, A.~Kolesnikov, D.~Weissenborn, X.~Zhai,
  T.~Unterthiner, M.~Dehghani, M.~Minderer, G.~Heigold, S.~Gelly, J.~Uszkoreit,
  and N.~Houlsby, ``An image is worth 16x16 words: Transformers for image
  recognition at scale,'' in {\em International Conference on Learning
  Representations}, 2021.

\bibitem{wav2vec2}
A.~Baevski, Y.~Zhou, A.~Mohamed, and M.~Auli, ``wav2vec 2.0: A framework for
  self-supervised learning of speech representations,'' {\em Advances in Neural
  Information Processing Systems}, vol.~33, pp.~12449--12460, 2020.

\bibitem{multichannel}
Y.~Zheng, Q.~Liu, E.~Chen, Y.~Ge, and J.~L. Zhao, ``Time series classification
  using multi-channels deep convolutional neural networks,'' in {\em
  International conference on web-age information management}, pp.~298--310,
  Springer, 2014.

\bibitem{Lusch2018}
B.~Lusch, J.~N. Kutz, and S.~L. Brunton, ``Deep learning for universal linear
  embeddings of nonlinear dynamics,'' {\em Nature Communications}, vol.~9,
  p.~4950, Nov. 2018.

\bibitem{Yeung2019}
E.~Yeung, S.~Kundu, and N.~Hodas, ``Learning deep neural network
  representations for {Koopman} operators of nonlinear dynamical systems,'' in
  {\em 2019 American Control Conference (ACC)}, pp.~4832--4839, 2019.

\bibitem{NIPS2017_3a835d32}
N.~Takeishi, Y.~Kawahara, and T.~Yairi, ``Learning {Koopman} invariant
  subspaces for dynamic mode decomposition,'' in {\em Advances in Neural
  Information Processing Systems} (I.~Guyon, U.~V. Luxburg, S.~Bengio,
  H.~Wallach, R.~Fergus, S.~Vishwanathan, and R.~Garnett, eds.), vol.~30,
  Curran Associates, Inc., 2017.

\bibitem{Drgona2022}
J.~Drgoňa, A.~Tuor, S.~Vasisht, and D.~Vrabie, ``Dissipative deep neural
  dynamical systems,'' {\em IEEE Open Journal of Control Systems}, vol.~1,
  pp.~100--112, 2022.

\bibitem{Skomski2021}
E.~Skomski, S.~Vasisht, C.~Wight, A.~Tuor, J.~Drgoňa, and D.~Vrabie,
  ``Constrained block nonlinear neural dynamical models,'' in {\em 2021
  American Control Conference (ACC)}, pp.~3993--4000, 2021.

\bibitem{NEURIPS2021_c9dd73f5}
J.~Drgona, S.~Mukherjee, J.~Zhang, F.~Liu, and M.~Halappanavar, ``On the
  stochastic stability of deep markov models,'' in {\em Advances in Neural
  Information Processing Systems} (M.~Ranzato, A.~Beygelzimer, Y.~Dauphin,
  P.~Liang, and J.~W. Vaughan, eds.), vol.~34, pp.~24033--24047, Curran
  Associates, Inc., 2021.

\bibitem{lusch2018deep}
B.~Lusch, J.~N. Kutz, and S.~L. Brunton, ``Deep learning for universal linear
  embeddings of nonlinear dynamics,'' {\em Nature communications}, vol.~9,
  no.~1, p.~4950, 2018.

\bibitem{takeishi2017learning}
N.~Takeishi, Y.~Kawahara, and T.~Yairi, ``Learning koopman invariant subspaces
  for dynamic mode decomposition,'' {\em Advances in neural information
  processing systems}, vol.~30, 2017.

\end{thebibliography}


\end{document}